\documentclass[sigconf]{acmart}

\usepackage{amsmath,amsfonts}
\usepackage{algorithmic}
\usepackage{algorithm}
\usepackage{array}
\usepackage{color}
\usepackage{textcomp}
\usepackage{stfloats}
\usepackage{url}
\usepackage{verbatim}
\usepackage{subfigure}

\usepackage{multirow} 
\usepackage{caption}  
\usepackage{booktabs} 
\usepackage{arydshln} 
\usepackage{bm}
\usepackage{paralist}
\usepackage{verbatimbox}
\usepackage{makecell} 

\usepackage{tcolorbox}
\usepackage{adjustbox}

\acmSubmissionID{1407}

\definecolor{mygreen}{RGB}{93,173,85}

\newcommand{\mysubsubsec}[1]{\noindent$\bullet$~\textbf{#1}}
\newcommand{\mybullet}[1]{\noindent$\bullet$~#1}

\def\eg{\emph{e.g.}}
\def\ie{\emph{i.e.}}
\definecolor{myred}{RGB}{252,66,70}
\definecolor{mypurple}{RGB}{234,236,255}
\definecolor{black}{RGB}{0,0,0}
\definecolor{myblue}{RGB}{221,247,255}

\newtcolorbox{mybox}[2][]{
width=\columnwidth,
colback =  mypurple,
colframe = white!75!white,, 
boxsep=0pt,left=10pt,right=10pt,top=3pt,bottom=5pt,
fontupper=\linespread{0.9}\selectfont,
title=#2,#1}

\newtcolorbox{mybluebox}[2][]{
width=\columnwidth,
colback =  myblue,
colframe = white!75!white,, 
boxsep=0pt,left=10pt,right=10pt,top=3pt,bottom=5pt,
fontupper=\linespread{0.9}\selectfont,
title=#2,#1}

\author{Leigang Qu}
\authornote{Both authors contributed equally to this research.}
\email{leigangqu@gmail.com}
\affiliation{%
  \institution{NExT Research Center, National University of Singapore}
  \country{}
}

\author{Shengqiong Wu}
\authornotemark[1]
\email{swu@u.nus.edu}
\affiliation{%
  \institution{NExT Research Center, National University of Singapore}
  \country{}
}

\author{Hao Fei}
\authornote{Hao Fei is the corresponding author. }
\email{haofei37@nus.edu.sg}
\affiliation{%
  \institution{NExT Research Center, National University of Singapore}
  \country{}
}

\author{Liqiang Nie}
\email{nieliqiang@gmail.com}
\affiliation{%
  \institution{Harbin Institute of Technology (Shenzhen)}
  \country{}
}

\author{Tat-Seng Chua}
\email{dcscts@nus.edu.sg}
\affiliation{%
  \institution{NExT Research Center, National University of Singapore}
  \country{}
}

\copyrightyear{2023}
\acmYear{2023}
\setcopyright{acmlicensed}\acmConference[MM '23]{Proceedings of the 31st
ACM International Conference on Multimedia}{October 29-November 3,
2023}{Ottawa, ON, Canada}
\acmBooktitle{Proceedings of the 31st ACM International Conference on
Multimedia (MM '23), October 29-November 3, 2023, Ottawa, ON, Canada}
\acmPrice{15.00}
\acmDOI{10.1145/3581783.3612012}
\acmISBN{979-8-4007-0108-5/23/10}

\begin{document}

\title{LayoutLLM-T2I: Eliciting Layout Guidance from LLM for Text-to-Image Generation}

\begin{abstract}
In the text-to-image generation field, recent remarkable progress in Stable Diffusion makes it possible to generate rich kinds of novel photorealistic images.
However, current models still face misalignment issues (\eg, problematic spatial relation understanding and numeration failure) in complex natural scenes, which impedes the high-faithfulness text-to-image generation. 
Although recent efforts have been made to improve controllability by giving fine-grained guidance (\eg, sketch and scribbles), this issue has not been fundamentally tackled since users have to provide such guidance information manually.  
In this work, we strive to synthesize high-fidelity images that are semantically aligned with a given textual prompt without any guidance. 
Toward this end, we propose a coarse-to-fine paradigm to achieve layout planning and image generation. 
Concretely, we first generate the coarse-grained layout conditioned on a given textual prompt via in-context learning based on Large Language Models.
Afterward, we propose a fine-grained object-interaction diffusion method to synthesize high-faithfulness images conditioned on the prompt and the automatically generated layout. 
Extensive experiments demonstrate that our proposed method outperforms the state-of-the-art models in terms of layout and image generation.
Our code and settings are available at \url{https://layoutllm-t2i.github.io/}.
\end{abstract}



\maketitle

\section{Introduction}\label{sec:intro}

\begin{figure}[t]
\includegraphics[width=0.87\columnwidth]{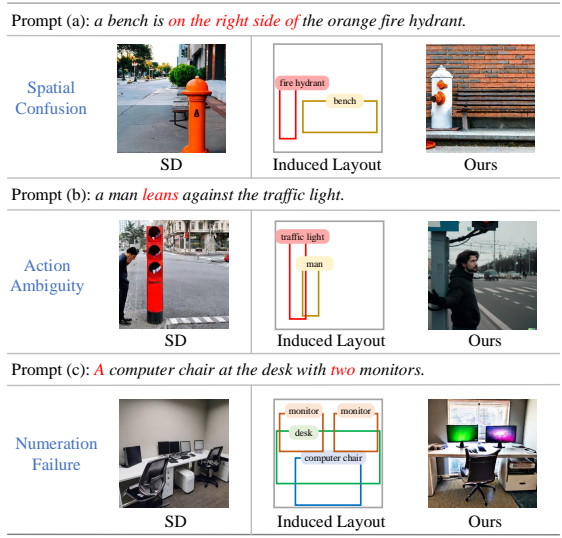}
\vspace{-2mm}
\caption{
Illustration of T2I task. 
Given the prompt, Stable Diffusion (SD) is subject to certain issues such as \emph{spatial confusion}, \emph{action ambiguilty} and \emph{numeration failure}. 
Our proposed model is able to synthesize high-faithfulness images by leveraging the automatically generated layouts. 
Numeration and relation terms in prompts are marked with \textcolor{myred}{red}. }
\label{fig:wrong_layout}
\vspace{-3mm}
\end{figure}

In the latest days, the topic of AI-Generated Content (AIGC) has made thrilling progress, such as DELL-E 2~\cite{ramesh2022hierarchical}, Stable Diffusion (SD) ~\cite{rombach2022high}, and ChatGPT~\cite{ouyang2022training}. 
As one of the representative generative AI themes, text-to-image generation (T2I) has received extensive attention from both academia and industry.
Given input language prompts, T2I aims to produce images that accurately reflect the desired contents as well as their semantic correlations.
Currently, the diffusion-based models have become the state-of-the-art (SoTA) T2I method, due to the preferable distribution coverage, a stationary training objective, and easy scalability \cite{HoJA20, DhariwalN21, rombach2022high}.
Despite the satisfactory performance achieved by recent SD-based models, 
synthesizing high-faithful images in complex scenes is still challenging \cite{abs-2205-11487,abs-2204-06125}.
In Figure \ref{fig:wrong_layout} we showcase several representative issues in current SD-based T2I,\footnote{Here we generate images using the official SD model with v1-4 checkpoint weights from \url{https://github.com/CompVis/stable-diffusion}.} such as \emph{problematic spatial relation understanding} and \emph{numeration failure}.

Diffusion models are competent in accurately rendering the visual objects by recognizing the explicit entity mentions of interest from prompt texts.
However, we argue that the key to high-faithfulness image synthesis, especially for complex scenes, also lies in the rigorous understanding of the underlying layout and delicate interactions between objects.\footnote{
We note that some recent SD-based methods, \eg, ControlNet \cite{abs-2302-05543}, in combination with additional human guidance can promisingly handle complex-scene T2I, while this work mainly considers fully automatic solutions without human efforts.
}
Intuitively, whenever we humans create a fine picture by the prompt instruction, we often follow a two-stage drawing process.
First, we pin down the general layout of the overall picture, \ie, sketching out all the objects as well as their relative semantic relations.
With the top-level design of the picture, we then complete all the necessary details.
In a nutshell, high-faithful image synthesis further requires the capability of high-level planning.
Inspired by such coarse-to-fine drawing intuition, in this work, we investigate endowing T2I models with scene layout planning abilities, such that the model is able to validly devise the coarse-grained architecture and the semantic structure before rendering the fine-grained details.

However, it is non-trivial to achieve high-faithfulness image synthesis via the above-mentioned coarse-to-fine framework, due to the following challenges. 1) \textbf{Layout Planning} requires abstract spatial imagination and analysis capabilities. The limited annotated layout data and intrinsic inductive bias make it difficult for existing diffusion methods~\cite{rombach2022high, NicholDRSMMSC22} to accurately and aesthetically generate layouts. Although notable efforts~\cite{li2023gligen, mou2023t2i, zhang2023adding} have been dedicated to synthesizing complex scenes by manually providing guidance information, these strategies suffer from weak flexibility and low efficiency since they heavily rely on extra labor-intensive guidance.
And 2) \textbf{Relation Modeling}, \eg, expressing high-level spatial and semantic relations, plays a pivotal role in understanding, imagining, and depicting complex scenes for T2I models, but it is still under-explored owing to the complex environments in real life.

\begin{figure*}[t]
	\includegraphics[width=0.88\textwidth]{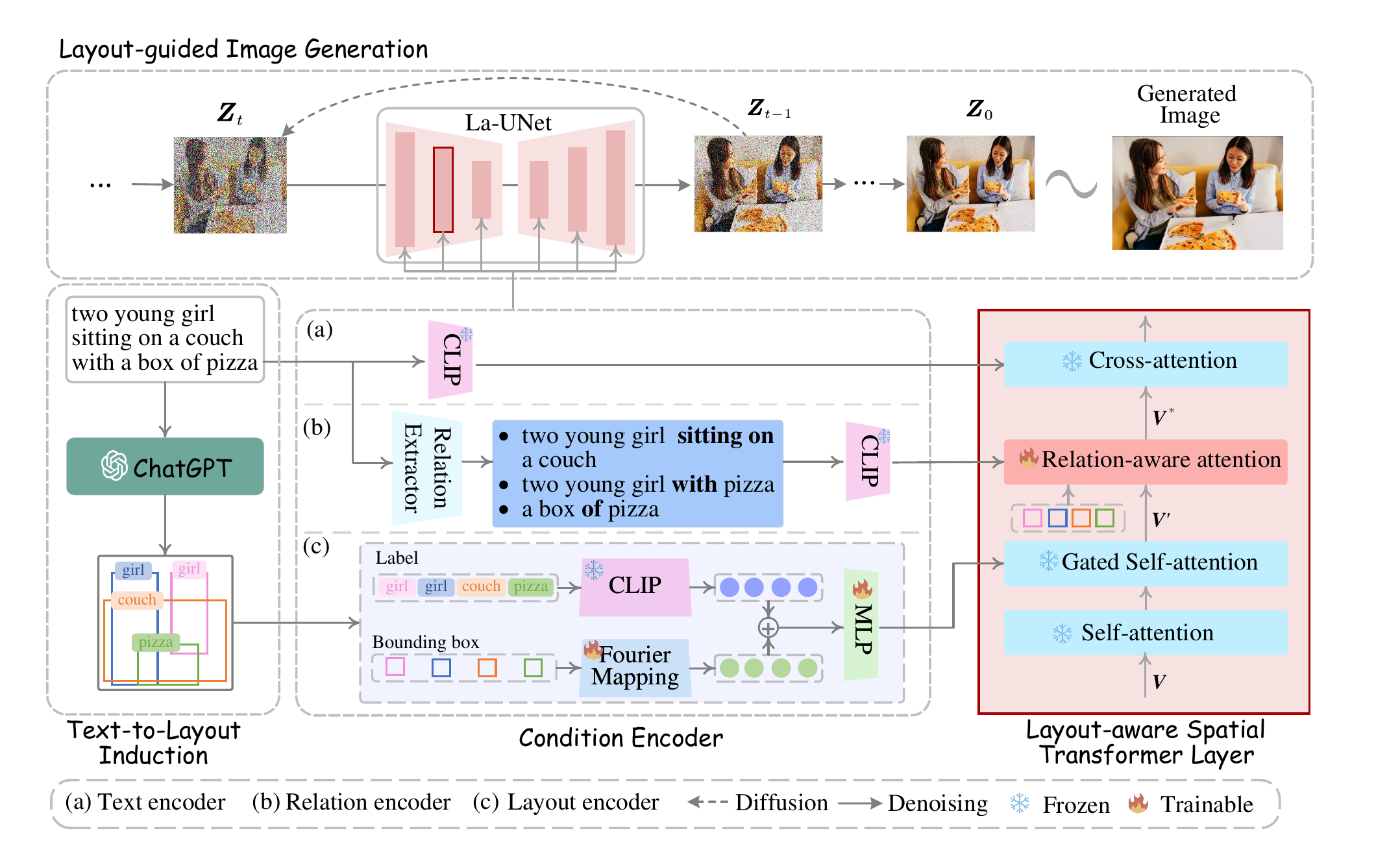}
	\vspace{-3ex}
	\caption{The overview of our framework. The proposed layout-guided diffusion model leverages a La-UNet, in which we first leverage ChatGPT to induce the layout from the given text prompt, and then a prompt encoder models the text prompt, the relation triplets (subject, relation, object) extracted from the text, and the induced layout separately. Finally, to efficiently integrate the layout information, we introduce a Layout-aware Spatial transformer based on the UNet.}
\label{fig:framework}
\vspace{-3ex}
\end{figure*}

Facing these two challenges, we propose an effective model by eliciting layout guidance from LLM for high-faithful T2I generation (\textbf{LayoutLLM-T2I}).
As shown in Figure \ref{fig:framework}, our framework comprises two main modules, including the text-to-layout induction and the layout-guided text-to-image generation.
In the first stage, we explore the scene understanding ability of large language models (LLMs), \eg, ChatGPT, for layout planning. 
To fully stimulate this ability, we design a feedback-based sampler learning mechanism, which is able to adaptively select informative examples for in-context learning, guided by layout-level and image-level feedback. 
During the second layout-guided text-to-image generation stage, 
based on the parameter-frozen SD model we devise a layout-aware adapter, in which the mentioned entities with well-organized layouts and their semantic relation information are injected into the backbone SD with fine-grained adequate interaction.
We perform extensive experiments on T2I benchmarks, where the proposed model achieves new SoTA results over existing methods, 
demonstrating the effectiveness of the layout guidance for diffusion-based T2I.
We also show that the proposed feedback-based sampler learning mechanism is beneficial in activating high-quality layout planning;
and the layout-guided adapter helps maintain effective layout feature integration for T2I synthesis.
In-depth experiments and analysis demonstrate that our method improves T2I generation, especially in complex-scene cases and zero-shot settings.

To summarize, our contributions are three-fold:
\begin{compactitem}
    
    \item To the best of our knowledge, it is the first work to investigate layout planning under complex natural scenes in the context of LLMs and Diffusion models.

    \item We propose a feedback-based sampler learning paradigm for layout generation and a layout-guided object interaction scheme for conditioned image synthesis.

    \item The proposed framework empirically pushes the current SoTA T2I performance, achieving high-faithfulness image synthesis in complex scenes.

\end{compactitem}

\section{Related Work}
Text-to-image generation, \emph{a.k.a.}, text-conditional image synthesis, has been the key research topic in the multimodal learning community.
There have been a number of efforts devoted to generating realistic and natural-looking images.
The generative adversarial networks (GANs)~\cite{GoodfellowPMXWOCB14,ReedAYLSL16} are a popular class of generative models that use a two-part network: a generator and a discriminator, while variational autoencoders (VAEs)~ \cite{KingmaW13} apply a probabilistic encoder-decoder architecture.
Recently, inspired by the application of auto-regressive models (ARMs) in text generation, numerous work has adopted ARMs to achieve impressive results for text-to-image  generation, such as DALL-E \cite{RameshPGGVRCS21}, CogView \cite{DingYHZZYLZSYT21}, and Pariti \cite{abs-2206-10789}.
Despite their success, existing T2I generation models still suffer from some weaknesses, such as training instability in GANs \cite{RombachBLEO22} and unidirectional bias in ARMs \cite{GuCBWZCYG22}. 
The diffusion models (DM) have currently emerged as the SoTA T2I approaches \cite{RombachBLEO22,NicholDRSMMSC22,abs-2205-11487}, due to the natural fit to inductive biases of image data, leading to remarkable synthesis quality. 
For example, \citet{RombachBLEO22} proposed a latent diffusion model that enables DM training on limited computational resources while retaining their quality and flexibility. 
\citet{NicholDRSMMSC22} presented GLIDE, an effective text-guidance strategy leading to photorealistic image generation and editing.

While most of the existing methods have secured satisfactory performance for the T2I \cite{ReedAYLSL16,BaoCWLH17}, generating high-fidelity images in complex scenes that faithfully reflect the original text prompt is still challenging~\cite{nie2022search, qu2021dynamic}.
In the realistic world, it is ubiquitous that user prompts come with complicated descriptions, \ie, multiple various objects in complex interrelation (such as spatial relations, action-based semantic relations, and numeric relations).
Correspondingly, prior efforts have been paid to model complex scenes, \eg, stacking multiple GANs \cite{ZhangXL17}, conditioning on scene graphs \cite{JohnsonGF18}, and introducing attentional generative networks \cite{XuZHZGH018}.
However, few attempts focused on enhancing the faithfulness for T2I.
\citet{abs-2212-05032} proposed to integrate the syntactic structure of the prompt sentence such that the key objects and the corresponding relations can be learned more correctly.
To strengthen the modeling of object spatial relations, additional segmentation features \cite{abs-2211-14305,GafniPASPT22}, or spatial conditioning \cite{abs-2208-12242,abs-2211-13752,abs-2302-08113,mou2023t2i,ReedAMTSL16,HinzHW19} are integrated into the visual synthesis process for achieving higher faithfulness.
In this work, we argue that the key to high-faithfulness T2I generation lies in the layout planning and comprehensive understanding of the underlying interactions between objects. 
We draw inspiration from human intuition and consider strengthening the image generation in complex scenes by taking advantage of the high-level layout features as guidance for high-fidelity diffusion-based T2I.

Previous research has demonstrated that modeling the high-level object layout information helps to capture the underlying abstract semantic relations and results in better vision generation \cite{JohnsonGF18,HongYCL18,VoS20}. 
Some works study the task of image synthesis from layout input (\ie, layout-to-image), where GANs \cite{SunW19,MaZS20,HeLYYSRX21} are employed.
Recently, diffusion models are adopted for layout-to-image and achieve more reliable image generation \cite{li2023gligen,abs-2302-08908,abs-2303-17189}. 
Different from these works, in this study, we focus on the T2I setting without giving any extra layout information, \ie, only textual prompts as input.
By eliciting the layout generation abilities from LLMs by a feedback-based sampler, we achieve high-quality layout label acquisition without relying on any human effort.
Besides, we devise an effective strategy to integrate layouts into the diffusion process.

\section{Preliminary on Latent Diffusion}

In this paper, we apply our method based on the open-sourced SD model \cite{RombachBLEO22}. 
SD employs a hierarchical VAE to operate the diffusion process in low-dimensional latent space, instead of operating in the image space, improving the computational efficiency.
Technically, an encoder $\mathcal{E}$ of VAE maps a given image $I$ into a spatial latent code $\bm{Z}$, \ie, $\bm{Z} = \mathcal{E}(I)$.
A diffusion model \cite{HoJA20} operates over the learned latent space to produce a denoised version of an input latent $\bm{Z}_t$ at each timestep $t$ conditioned on addition input.
In a text-to-image scenario, this additional input is typically a text encoded by a pre-trained CLIP text encoder $\tau_\theta$~\cite{RadfordKHRGASAM21}.
During the training process, at each timestep $t$, the denoising network $\varepsilon_\theta$ is optimized to remove the noise $\varepsilon$ added to the latent code $\bm{Z}$, given the noised latent $\bm{Z}_t$, the timestep $t$, and the conditioning text $y$:
\begin{equation}
\label{eq:diffusion}
    \mathcal{L} = \mathbb{E}_{Z  \sim \mathcal{E}(I), y,\varepsilon \sim \mathcal{N}(0, 1),t} \Big[\Vert\varepsilon - \varepsilon_\theta (\bm{Z}_t, t, \tau_\theta(y))\Vert_2^2\Big] \,.
\end{equation}
Here, $\varepsilon_\theta$ is often implemented with a UNet \cite{RonnebergerFB15} consisting of convolution, self-attention, and cross-attention layers.

At inference time, a sampling process is performed to iteratively denoise with $\bm{Z}_T \in \mathcal{N}(0, I)$ as the start.
Specifically, at each denoising step $t = 1, \cdots, T$, $\bm{Z}_{t-1}$ is obtained by denoise $\bm{Z}_t$ conditioned on the text prompt $y$. 
After the final denoising step, $\bm{Z}_0$ will be mapped back to the original image space, generating an image by a decoder of VAE, $I^{'} = \mathcal{D}(\bm{Z}_0)$.

\section{Methodology}

In Figure \ref{fig:framework}, we illustrate the overall architecture of the proposed layout-guided diffusion model, consisting of two modules. 
First, the text-to-layout induction module (Section~\ref{sec:layout_generation}) infers a coarse-grained layout via an LLM conditioned on the given textual prompt. Combining the prompt and the generated layout, the layout-guided image generation module (Section~\ref{sec:Layout-guided-T2I}) synthesizes the final image.
In what follows, we will delve into these two modules.

\begin{figure}[t]
\includegraphics[width=0.75\columnwidth]{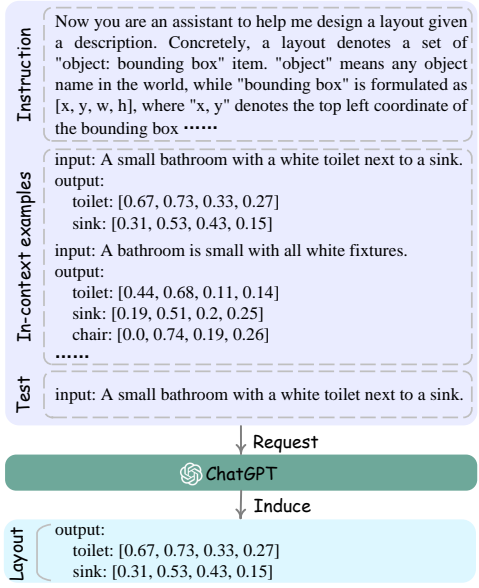}
\caption{Schematic illustration of layout generation.}
\label{fig:layout_generation}
\vspace{-5mm}
\end{figure}

\subsection{Text-to-Layout Induction}
\label{sec:layout_generation}
Recent years have witnessed the tremendous potential of LLMs~\cite{touvron2023llama, abs-2303-08774, abs-2204-02311}. Benefiting from the large corpus and ample computing resources, they achieve outstanding performance in most natural language processing (NLP) tasks, especially under the challenging zero-shot or few-shot settings~\cite{abs-2303-08774}.  
The impressive success of LLMs in NLP illustrates the multifaceted abilities of LLMs.
Inspired by it, we aim to excavate the spatial imagination, semantic relation, and numeration understanding abilities of LLMs toward layout planning and facilitate the text-to-image generation task.

Concretely, we resort to in-context learning (ICL) \cite{abs-2201-11903} to activate LLMs for layout generation. 
Typically, ICL employs a natural language prompt that includes a task description (\texttt{Instruction}), a few examples (\texttt{in-context examples}) selected from the training dataset as demonstrations, and a test instance (\texttt{Test}), as depicted in Figure \ref{fig:layout_generation}. 
Previous studies have shown that the effectiveness of ICL is highly influenced by the design of demonstrations \cite{MinLHALHZ22,LuBM0S22,ZhaoWFK021}. 
Therefore, it is essential to select a subset of examples that can effectively leverage the ICL capability of LLMs. 
To tackle this issue, we devise an adaptive sampler based on layout-level and image-level feedback to select examples in a reinforcement learning framework.
This framework mainly consists of three parts, \ie, policy network, reward, and optimization.

\mysubsubsec{Policy Network.}  
We randomly sample instances from the training set to form a candidate set $\mathcal{C}$. Given a text $y_i$,  we aim to select $K$ suitable in-context examples $\mathcal{C}_i = \{c_i^k | k = 1, 2, ..., K\} \subset \mathcal{C}$. The selection is modeled by a policy network parameterized by $\psi$:
\begin{equation}
    c_i^k \sim \pi_\psi (c_i | y_i), 
\end{equation}
where $c_i^k \in \mathcal{C}$ is independently sampled from the candidate set. In practice, the policy is implemented as, 
\begin{equation}\label{eqn:policy_implemented}
    \pi_\psi (c_i | y_i) = \frac{\exp (f(y[c_i]) \cdot f(y_i))}{\sum_{{c'} \in \mathcal{C}} \exp (f(y[c']) \cdot f(y_i))},
\end{equation}
where $y[c_i]$ denotes the text with respect to the candidate $c_i$. $f (\cdot)$ acts as a mapping function that transforms a text into a latent layout embedding. In this latent space, two sentences describing similar layouts will be mapped close to each other. 

Combining the given text, the selected in-context examples, and the instruction as the prompt, we obtain the layout $\hat{b}_i$ from an LLM:
\begin{equation}
    \hat{b}_i = {\rm LLM}(y_i, \mathcal{C}_i).
\end{equation}

\begin{figure}[t]
	\includegraphics[width=0.85\columnwidth]{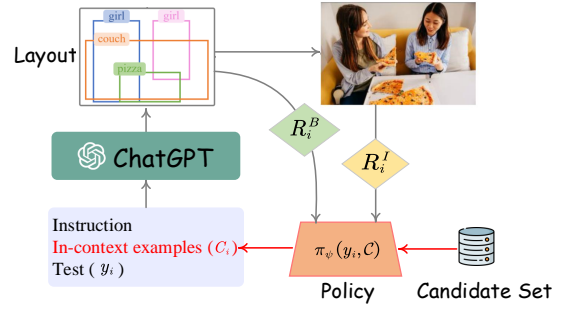}
	\caption{The layout-image feedback module consists of a policy network $\pi_\psi (y_i, C)$ and two rewards $R_i^I$ and $R_i^B$. Guided by these two rewards, the policy learns to sample informative training data instances as the context fed into LLMs to activate the layout planning abilities. }
\label{fig:icl}
\end{figure}

\mysubsubsec{Reward.}
As discussed in Section~\ref{sec:intro}, layouts play a key role in text-to-image generation without other fine-grained guidance. Meanwhile, the final aim is to generate a reasonable and aesthetic image to satisfy user intention. With these two aspects in consideration, based on the generated layout $\hat{b}_i$, we define the total reward as:
\begin{equation}
    R(\hat{b}_i | y_i) = R^B_i + R^I_i, 
\end{equation}
where $R^B_i$ and $R^I_i$ denote the layout reward and image reward, respectively. Specifically, they are calculated by: 
\begin{equation}
\begin{split}
    R_i^B &= {\rm mIoU} (\hat{b}_i, b_i), \\
    R_i^I &= {\rm Sim} (\hat{x}_i, x_i, y_i) + {\rm Aes} (\hat{x}_i), 
\end{split}
\end{equation}
where ${\rm mIoU}(\hat{b}_i, b_i)$ refers to the maximum intersect over union~\cite{KikuchiSOY21} between the induced layout $\hat{b}_i$ and the ground-truth layout $b_i$, measuring the layout similarity in the spatial dimension. 
Besides, ${\rm Sim} (\hat{x}_i, x_i, y_i)$ denotes the CLIP~\cite{RadfordKHRGASAM21} similarity of the generated image $\hat{x}_i$ from $\hat{b}_i$ to the ground-truth one $x_i$ and the given text $y_i$, respevtively. 
Concretely, we employ both intra-modal (image-to-image) and cross-modal (image-to-text) similarities, \ie, ${\rm Sim} (\hat{x}_i, x_i, y_i)$ = ${\rm CLIP}_{I \leftrightarrow I} (\hat{x}_i, x_i) + {\rm CLIP}_{I \leftrightarrow T} (\hat{x}_i, y_i)$. 
In addition to the semantic alignment, we also consider another aspect, \ie, \textit{aesthetics}, to measure the image generation quality.
In detail, we adopt the aesthetic predictor\footnote{\url{https://github.com/christophschuhmann/improved-aesthetic-predictor}} trained on the LAION dataset~\cite{schuhmann2022laion} to calculate the aesthetic score ${\rm Aes} (\hat{x}_i)$.

\mysubsubsec{Optimization.}
To optimize the policy network, we first carry out Monte Carlo Sampling~\cite{shapiro2003monte} to estimate the expected reward:
\begin{equation}
    \mathbb{E}_{c_i \sim \pi_\psi (c_i | y_i)}[R({\rm LLM}(y_i, \mathcal{C}_i))] \approx \frac{1}{N} \sum_{i=1}^{N} R({\rm LLM}(y_i, \mathcal{C}_i)), 
\end{equation}
in which $N$ denotes the batch size. And then we perform optimization using the REINFORCE policy gradient algorithm~\cite{williams1992simple}:
\begin{equation}
\begin{split}
    & \bigtriangledown \mathbb{E}_{c_i \sim \pi_\psi (c_i | y_i)}[R({\rm LLM}(y_i, \mathcal{C}_i))] \\
    &= \mathbb{E}_{c_i \sim \pi_\psi (c_i | y_i)} \bigtriangledown_\psi \log (\pi_\psi(c_i | y_i)) R({\rm LLM}(y_i, \mathcal{C}_i)) \\
    &\approx \frac{1}{N} \sum_{i=1}^N \bigtriangledown_\psi \log (\pi_\psi(c_i | y_i)) R({\rm LLM}(y_i, \mathcal{C}_i)).
\end{split}
\end{equation}
By maximizing the expected reward, the policy network learns to select those in-context examples which motivate the LLM to generate a reasonable and aesthetic layout. Meanwhile, the induced layout could guide the image generation model to synthesize a high-quality and high-faithfulness image.

\subsection{Layout-guided Image Generation} 
\label{sec:Layout-guided-T2I}
In the above coarse-grained layout planning process, we activate an LLM to generate reasonable and aesthetic layouts. However, an accurate layout does not guarantee high-faithfulness image generation, since the same layout can induce multiple images with different semantics. 
For example, given the two prompts, ``A man walks towards a traffic light'' and ``A man looks at the traffic lights'',  two similar spatial arrangements could be obtained, where a man is on the left side of the image and a traffic light is on the right side. In light of this,  
it is essential to consider semantic relation modeling and scene understanding during the image generation process.
Toward this end, we endow such capabilities to the diffusion model via relation-aware object interaction.

\mysubsubsec{Condition Encoder.}
To encode the text prompt $y$, we leverage the pre-trained  CLIP~\cite{RadfordKHRGASAM21} to yield a feature sequence $\bm{H}^{y}$.
Furthermore, resorting to scene graph parser\footnote{https://github.com/vacancy/SceneGraphParser}, we capture explicit semantic relations by extracting object-predicate-object phrases $R = \{r_1, \cdots, r_m\}$, and then represent them as: 
\begin{equation}
    \bm{H}^{r} = \{\text{CLIP}(r_1), \cdots, \text{CLIP}(r_m)\}.
\end{equation}

After the layout induction presented in Section~\ref{sec:layout_generation}, we obtain the layout $B= \{(b_1, l_1), \cdots, (b_n, l_n)\}$ in which $l_i$ represent the object textual label of the bounding box $b_i$. 
Afterwards, we encode a bounding box coordinated with \texttt{Fourier}~\cite{TancikSMFRSRBN20} mapping. 
As shown in Figure \ref{fig:framework}, we concatenate label features and  bounding box features, and feed them into a multi-layer perception (MLP):
\begin{equation}
    \bm{H}^b = \text{MLP}(\text{Fourier}(\{b_1, ..., b_n\});\text{CLIP}(\{l_1, ..., l_n\})), 
\end{equation}
where $\bm{H}^{b} = \{\bm{h}_1^b, \cdots, \bm{h}_n^{b}\}$ denotes the layout feature sequence.

\mysubsubsec{Relation-aware Image Generation.}
Existing models~\cite{li2023gligen, mou2023t2i, zhang2023adding} have demonstrated the potential of SD to generate high-quality images based on layout information offered by users. 
In this paper, based on GLIGEN~\cite{li2023gligen}, we present the relation-aware image generation module. 
In GLIGEN, two attention layers are frozen in the original Transformer block of SD, and an extra gated self-attention layer is added as an adapter to model the cross-modal interaction between intermediate visual features $\bm{V}$ and layout features $\bm{H}^{b}$:
\begin{equation}\label{eq:gated_attn}
    \bm{V}' = \bm{V} + \beta \cdot \text{Tanh}(\gamma) \cdot \text{TS} (\text{SelfAttn}([\bm{V},\bm{H}^{b}])),
\end{equation}
where $\text{TS}(\cdot)$ is a token selection operation that considers visual tokens only and $\gamma$ is a learnable scalar. 
$\beta$ acts as a hyperparameter to balance quality and controllability. 

Though the self-attention operation over the combination of $\bm{V}$ and $\bm{H}^b$ encourages the interaction between layout, text, and image tokens, the intact visual object, as well as relations, are not considered.
Therefore, we first select visual objects\footnote{Note that $\bm{V}$ denotes visual tokens. Consequently, an intact object may be divided into multiple tokens, and one token may also consist of several objects.} and obtain their feature maps according to the bounding box:
\begin{equation}
    \bm{o}_i = \sum \bm{M}_i \odot \bm{V}', 
\end{equation}
where $\bm{M}_i$ denotes the mask induced from the bounding box $b_i$.  
After obtaining all the object features $\bm{O}=[\bm{o}_1; \cdots; \bm{o}_n]$, we apply the across-attention to integrate relation information into the model:
\begin{equation}
\label{eq:rela_attn}
    \bm{V}^* = \bm{V}' + \text{CrossAttn}(\bm{O}, \bm{H}^{r}, \bm{H}^{r}), 
\end{equation}
Note that Eq.\eqref{eq:rela_attn} is injected in between the gated self-attention layer and the cross-attention layer as shown in Figure \ref{fig:framework}.

\subsection{Optimization}
\label{sec:learning_procedure}

We adopt the pre-trained diffusion model such that layout information can be injected while all the original components remain intact.
By denoting the new parameters as $\theta^{'}$, we use the original denoising objective as in Eq.\eqref{eq:diffusion} for the model's continual learning, based on the text prompt $y$ and layout instructions $B$.
Finally, the generation process can be optimized via:
\begin{equation}
    \mathcal{L} = \mathbb{E}_{\bm{Z} \sim \mathcal{E}(I), y,\varepsilon \sim \mathcal{N}(0, 1),t} \Big[||\varepsilon - \varepsilon_{\theta,\theta^{'}} (\bm{Z}_t, t, y, B)||_2^2\Big] \,.
\end{equation}

\section{Experiments}
In this section, we carried out extensive experiments on COCO2014, the widely used benchmark dataset in vision understanding and generation, to answer the following research questions:

\mysubsubsec{RQ1}: How does the proposed method perform in the challenging layout planning and high-faithfulness image synthesis compared with state-of-the-art baselines?\\
\mysubsubsec{RQ2}: How does each component of the proposed method affect the performance of layout generation and image synthesis?\\
\mysubsubsec{RQ3}: How are the authenticity and rationality of the generated layouts and images?

\begin{table}[!t]
 \setlength{\tabcolsep}{1.15mm}
\caption{
    Overall performance comparison on the constructed test set of COCO 2014 for text-to-layout generation and text-to-image generation. The best results are highlighted in \textbf{bold}. 
    }
    \scalebox{0.80}{
        \vspace{-2mm}
            \centering
            \renewcommand\arraystretch{1.3}
            \begin{tabular}{l| ccc|  ccc}
            \Xhline{4\arrayrulewidth}
            \multicolumn{1}{l|}{\multirow{2}{*}{Methods}}& \multicolumn{3}{c|}{Layout} & \multicolumn{3}{c}{Image} \\ 
             & FID$\downarrow$ & mIoU$\uparrow$ & LaySim$\uparrow$ & FID$\downarrow$ & Sim (I-T)$\uparrow$ & Sim (I-I)$\uparrow$\\ 
            \hline\hline
            LayoutTrans~\cite{GuptaLA0MS21} & 31.51  & 0.95 & 0.41 & 85.75  & 19.63  & 53.94  \\
            MaskGIT~\cite{ChangZJLF22} &   87.09   & 6.46 & 3.73 & \underline{67.72}  & \underline{31.65}  & \underline{63.78}  \\
            BLT~\cite{KongJCZHGE22}  & 110.33  & 3.81 & 2.64 & 71.33  & 29.04  & 61.68  \\
            VQDiffusion~\cite{GuCBWZCYG22} & \underline{29.44}  & 6.98 & \underline{4.67} & \textbf{66.58}  & 31.49  & 63.04  \\
            LayoutDM~\cite{abs-2303-08137} & \textbf{23.69}  & \underline{7.86} & 4.50 & 68.38  & 29.55  & 62.04  \\
            \cdashline{1-7}
            \textbf{Ours (two-shot)} & 80.73  & \textbf{10.62} & \textbf{6.86}  & 71.02  & \textbf{53.38}  & \textbf{67.89}  \\
            \bottomrule
            \end{tabular}
    }
\label{table:overall_performance}
\vspace{-2mm}
\end{table}

\begin{table*}[!t]
\caption{
    Quantitative comparison for text-to-layout generation on the constructed test set of COCO 2014. `Numerical', `Spatial', and `Semantic' denote that the captions include numerical descriptions, spatial relationships, and semantic actions, respectively. `Mixed' denotes that the given prompts include multiple relations or numerical descriptions. `Null' refers to that there are not any explicit relation keywords in prompts, which requires more abstract reasoning and imagination abilities. }
    \vspace{-2mm}
    \centering
    \scalebox{0.82}{
        \setlength\tabcolsep{3.8pt}
        \renewcommand\arraystretch{1.2}
        \begin{tabular}{l| cc| cc| cc| cc| cc}
        \Xhline{4\arrayrulewidth}
        \multicolumn{1}{l|}{\multirow{2}{*}{Methods}}& \multicolumn{2}{c|}{Numerical} & \multicolumn{2}{c|}{Spatial} & \multicolumn{2}{c|}{Semantic} & \multicolumn{2}{c|}{Mixed} & \multicolumn{2}{c}{Null} \\ 
        
        &  mIoU$\uparrow$ & LaySim$\uparrow$ & mIoU$\uparrow$ & LaySim$\uparrow$ & mIoU$\uparrow$ & LaySim$\uparrow$  & mIoU$\uparrow$ & LaySim$\uparrow$ & mIoU$\uparrow$ & LaySim$\uparrow$\\ 
        \hline\hline
        LayoutTrans~\cite{GuptaLA0MS21} &  1.02 & 0.39  &  0.94 & 0.13   &  \underline{2.21} & 0.67   & 0.99 & 0.28 & \underline{0.80} & 0.21\\
        MaskGIT~\cite{ChangZJLF22} &  \underline{5.86} & 3.01   &  0.71 & 3.77 & 1.05 & 4.74  & 7.87 & 4.85   & 0.28 & \underline{3.98} \\ 
        BLT~\cite{KongJCZHGE22} &  3.24 & 2.38  & 0.39 & 2.17 & 1.17 & 3.25   & 4.56 & 3.12   & 0.31 & 2.41 \\ 
        VQDiffusion~\cite{GuCBWZCYG22}  &  5.63 &  \underline{3.44}  &  1.21 & 5.00  & 1.52 & 4.46   & 7.95 & 5.28  & 0.22 & 3.77\\ 
        LayoutDM~\cite{abs-2303-08137} & 5.80 & 2.83 &  0.84 &  \underline{5.48} &  1.73 & \underline{4.85}   & \underline{8.68} & \underline{6.41}  & 0.79 & 3.73  \\
        \cdashline{1-11}
        \textbf{Ours (two-shot)} &  \textbf{10.69}  & \textbf{6.88}    & \textbf{10.22}  & \textbf{6.42}    & \textbf{10.30}  & \textbf{7.39}    & \textbf{12.08}  & \textbf{6.70}   & \textbf{9.94}  & \textbf{6.88}  \\
        \bottomrule
        \end{tabular}
        }
\label{tab:layout_five}
\end{table*}

\subsection{Experimental Settings}
\subsubsection{\bf Datasets}
We conduct experiments on COCO \cite{LinMBHPRDZ14}, which contains 82,783 training images and 40,504 test images over 80 semantic classes, where each image is associated with instance-wise annotations (\ie, object bounding boxes and segmentation masks) and 5 text descriptions.
We split the training data into 95\% for training and 5\% for validation.

To thoroughly evaluate the layout planning and relation understanding abilities, we re-organize the raw test set and construct a new one. Concretely, we first pre-processed captions by means of NLP tools~\cite{bird2009natural} and then select those samples which require specific layout planning capabilities. Finally, we obtain a new test set including five categories, \ie, numerical, spatial, semantic, mixed, and null. 
Appendix \S\ref{appendix:testset construction} gives more details of this part.

\subsubsection{\bf Evaluation Metrics}
For quantitative evaluation, we employ the following metrics with respect to layout generation and image generation. 1) \textit{Layout Evaluation}: Following prior work~\cite{abs-2303-08137, KongJCZHGE22}, we adopt layout-level {Fr\'echet Inception Distance} (FID)~\cite{HeuselRUNH17}, Maximum IoU (mIoU)~\cite{KikuchiSOY21}, and Layout Similarity (LaySim)~\cite{abs-2303-08137} to assess the layout induction performance.
2) \textit{Image Evaluation}: We use image-level FID, cross-modal similarities (Sim(I-T)) and intra-modal ones (Sim(I-I)) to evaluate image generation quality. Refer to the Appendix \S\ref{apendix:detailed evaluation} section for more details.

\begin{table*}[!t]
\caption{
    Quantitative comparison for layout-guided text-to-image generation on the constructed test set of COCO 2014. 
    }
    \vspace{-2mm}
    \centering
    \scalebox{0.80}{
        \setlength\tabcolsep{3.8pt}
        \renewcommand\arraystretch{1.2}
        \begin{tabular}{l| cc| cc| cc| cc| cc}
        \Xhline{4\arrayrulewidth}
        
        & \multicolumn{2}{c|}{Numerical} & \multicolumn{2}{c|}{Spatial} & \multicolumn{2}{c|}{Semantic} & \multicolumn{2}{c|}{Mixed} & \multicolumn{2}{c}{Null} \\ 
        \multicolumn{1}{l|}{\multirow{-2}{*}{Methods}} & Sim (I-T)$\uparrow$ & Sim (I-I)$\uparrow$ & Sim (I-T)$\uparrow$ & Sim (I-I)$\uparrow$ & Sim (I-T)$\uparrow$ & Sim (I-I)$\uparrow$ & Sim (I-T)$\uparrow$ & Sim (I-I)$\uparrow$ & Sim (I-T)$\uparrow$ & Sim (I-I)$\uparrow$\\ 
        \hline\hline
        LayoutTrans~\cite{GuptaLA0MS21}  & 15.90  & 51.72   & 17.14  & 52.75   & 21.89  & 55.20   & 22.27  & 56.91   & 20.24 & 52.82 \\
        MaskGIT~\cite{ChangZJLF22}  & \underline{29.57}  & \underline{63.97}   & \underline{31.69}  & \underline{63.05}  & 32.91  & \underline{64.90}   & 29.64  & 63.62  & \underline{33.85} & \underline{63.39} \\
        BLT~\cite{KongJCZHGE22}  & 28.31  & 62.04   & 27.95  & 60.98  & 33.17  & 63.17    & 26.74  & 61.89   & 28.71 & 60.43 \\
        VQDiffusion~\cite{ChangZJLF22}  & 24.09  & 61.34  & 29.78  & 62.76   & \underline{36.46}  & 64.74  & \underline{32.02}  & \underline{63.63}   & 33.45 & 62.40 \\
        LayoutDM~\cite{abs-2303-08137}  & 25.98  & 61.60   & 31.75  & 62.20  & 31.36  & 63.75  & 28.04  & 61.69  & 29.75 & 60.84 \\
        \cdashline{1-11}
        \textbf{Ours (two-shot)} & \textbf{56.25}  & \textbf{68.10}  & \textbf{55.51}  & \textbf{67.92}   & \textbf{46.76}  & \textbf{67.88}   & \textbf{58.96}  & \textbf{68.87}   & \textbf{50.39}  & \textbf{67.19}  \\
        \bottomrule
        \end{tabular}
        }
\label{tab:img_performance}
    \vspace{-1mm}
\end{table*}

\subsubsection{\bf Baselines}
To evaluate the effectiveness of the proposed method, we compare it with the following layout generation baselines:
\textit{LayoutTrans}~\cite{GuptaLA0MS21} is a self-attention framework capturing the contextual relationships and generating layouts of graphical elements.
\textit{BLT}~\cite{KongJCZHGE22} introduce a bidirectional layout transformer to empower the transformer-based models.
\textit{MaskGIT}~\cite{ChangZJLF22} propose to learn a bidirectional
transformer by masked visual token prediction.
\textit{VQDiffusion}~\cite{GuCBWZCYG22} is based on a VQ-VAE whose latent space is modeled by a conditional variant of the recently developed discrete diffusion model.
\textit{LayoutDM}~\cite{abs-2303-08137} adopt the VQDiffusion to handle the structure layout data in a discrete representation.

\subsubsection{\bf Implementation Details}
Based on the pre-trained GLIGEN~\cite{li2023gligen}, we add extra relation-aware layers to model semantic relations and perform continual learning.
We take the \textit{gpt-3.5-turbo} model via OpenAI API\footnote{\url{https://platform.openai.com/docs/models/gpt-3-5}} as our LLM. Under the few-shot setting, we randomly sample 64 instances for training, and 32 instances to form the candidate set. Besides, we set 2 as the shot number by default. During the optimization phase, the total number of epochs, the batch size, and the initial learning rate are set to 80, 8, and $2 \times 10^{-4}$, respectively. One can refer to Appendix \S\ref{appendix:detailed settings} for more details.

\subsection{Performance Comparison (RQ1)}
To justify the overall effectiveness of the proposed model, we carry out extensive experiments to evaluate the quality of generated layouts and images. As shown in Table~\ref{table:overall_performance}, we can see that the proposed method substantially outperforms the compared baselines, achieving state-of-the-art results, especially on the pair-wise relevance metrics. Next, to further assess the validity and superiority of the proposed model, we evaluate the proposed method from five aspects with respect to layout planning and image generation. 

\subsubsection{\bf Text-to-Layout Generation}
First, we evaluate the text-to-layout generation capability in terms of numerical, spatial, and semantic modeling, as shown in Table~\ref{tab:layout_five}. The results show that the proposed method achieves the best performance under most evaluation metrics, \eg, mIoU and LaySim, substantially surpassing the compared baselines. 
To further explore how the proposed approach performs on complex scenes and abstract prompts, we perform another two groups of experiments, \ie, ``Mixed'' and ``Null''. As for complex scenarios with mixed relations and abstract prompts without any explicit relations, the proposed model remarkably surpasses all the existing baselines. 
These results demonstrate the superiority of the proposed layout-guided text-to-image generation model.

\subsubsection{\bf Layout-guided Text-to-Image Generation}
Based on the layouts generated by different methods, we employ the proposed layout-guided T2I generation model to synthesize images on the constructed test set of COCO 2014 in real-world scenes. 
From the results shown in Table~\ref{tab:img_performance}, we have the following observations:

\mybullet{The auto-regressive model LayoutTrans performs worst compared with other methods in all the evaluation metrics for image generation, indicating the limitation of the traditional auto-regressive paradigm for the layout-guided image generation task. 
}

\mybullet{
LayoutDM, VQDiffusion, BLT, and MaskedGIT gain similar performance in text-to-image generation, and this similarity is a direct reflection of their comparable layout generation capabilities.
 }

\mybullet{The proposed method exhibits a substantial performance advantage over the existing baselines, as evidenced by a remarkable improvement observed in layout induction. 
This outcome further suggests that LLMs possess spatial and relational reasoning capabilities, which can be effectively harnessed for the demanding task of layout-based image generation.}

\begin{figure}[t]
	\subfigure[In-context Example Sampling]{
		\label{fig:abl_rl}
		\includegraphics[width=0.24\textwidth]{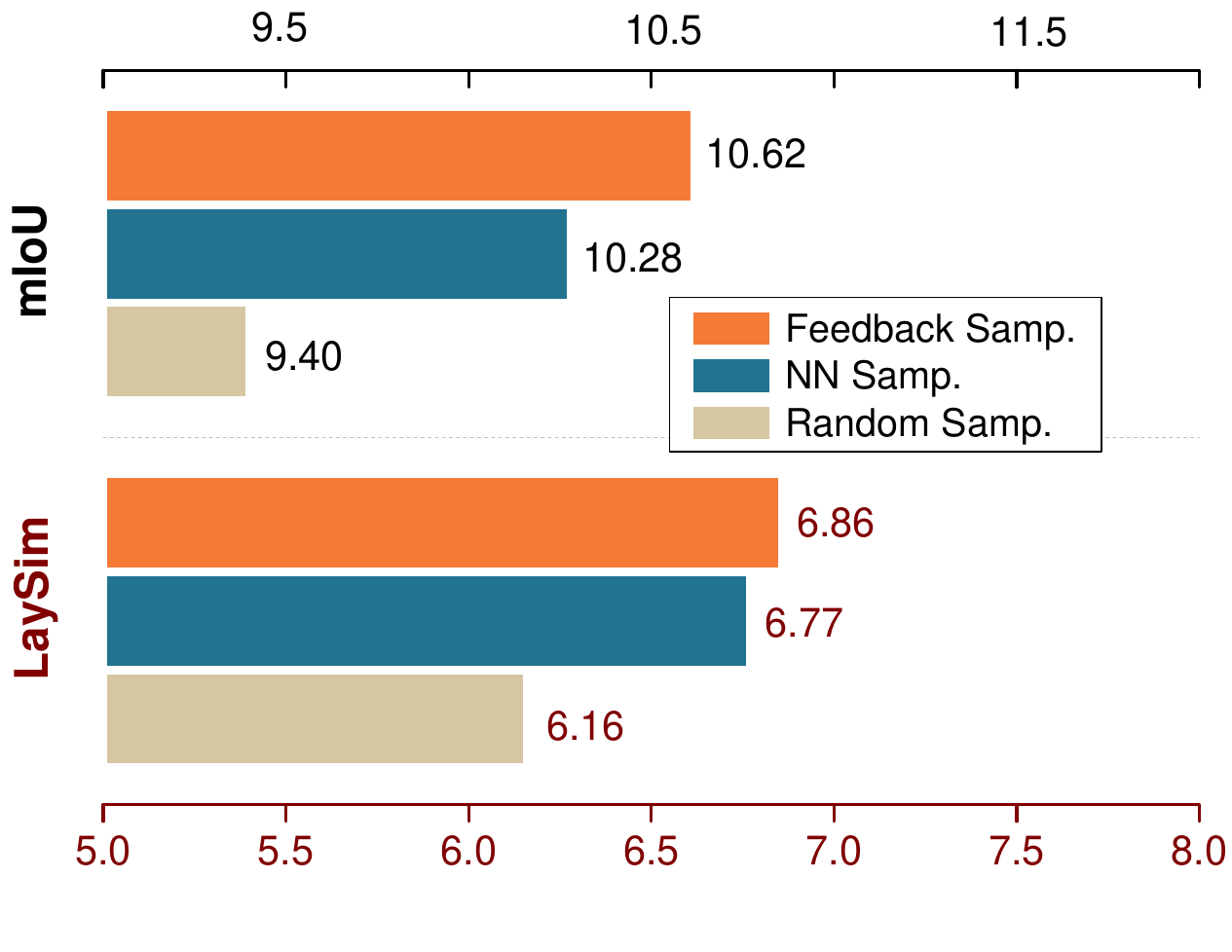}}	
	\hspace{-2ex}
	\subfigure[Shot number]{
		\label{fig:sensitive_shot}
		\includegraphics[width=0.23 \textwidth]{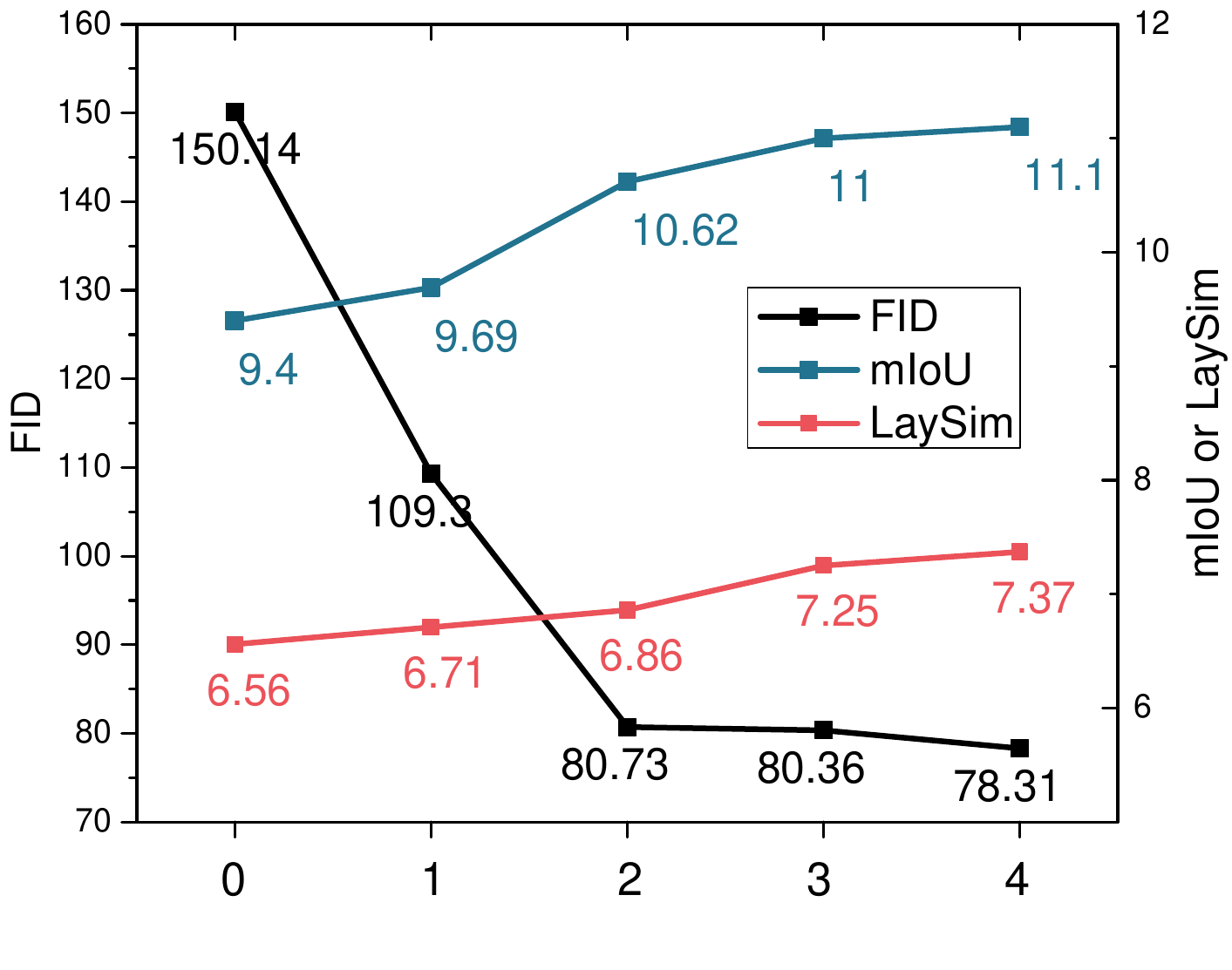}}
	\vspace{-3.0ex}
	\caption{Comparison of the (a) in-context example sampling strategies and (b) shot numbers for layout performance. Random and NN Samp. denote random sampling and the nearest neighbor sampling, respectively. }
	\vspace{-2.0ex}
\end{figure}

\begin{figure}[t]
	\includegraphics[width=0.35\textwidth]{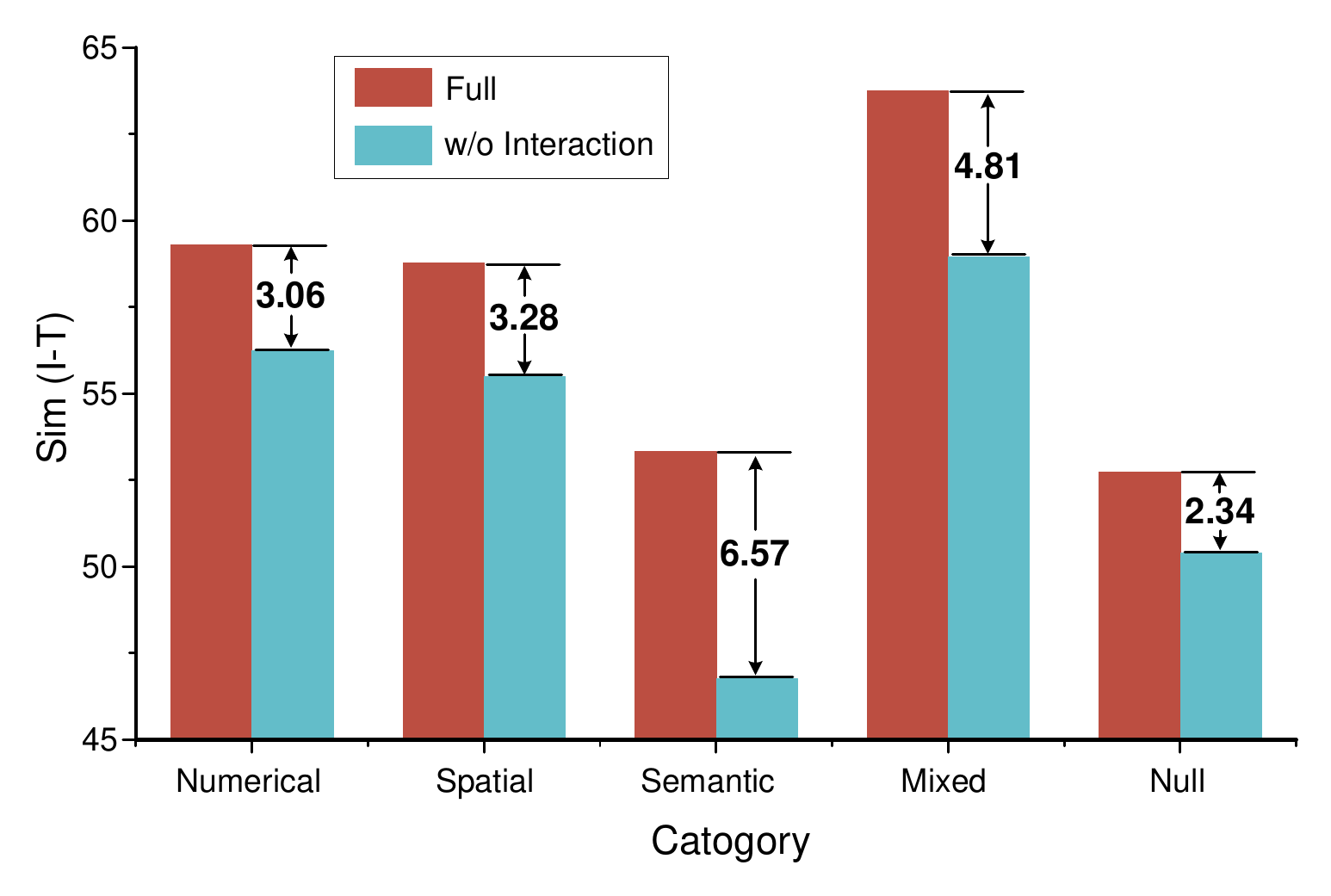}
	\vspace{-2ex}
	\caption{Ablation study on the proposed interaction-based relation-aware image generation module. We remove this module (w/o Interaction) and compare it with the Full model on the cross-modal alignment metric, \ie, Sim (I-T). }
	\label{fig:abl_interaction}
	\vspace{-3ex}
\end{figure}

\subsection{In-depth Analysis (RQ2 \& RQ3)}

\subsubsection{\bf Ablation Study}
Here we present model ablations to ascertain the efficacy of each part of the proposed method, including the feedback-based sampling strategy, the shot number of in-context examples, and the relation-aware image generation module, as elaborated subsequently.

\mysubsubsec{Impact of Feedback-based Sampling.}
Previous studies~\cite{abs-2201-11903, ZhangFT22} have indicated that the activation of certain abilities of LLMs necessitates appropriate examples combined with corresponding questions for in-context learning.
To facilitate the spatial comprehension, language-layout alignment, and layout planning abilities of LLMs, we propose the feedback-based sampling strategy.
To assess its effectiveness and investigate the impact of various sampling strategies on LLMs, we design two additional variants: 1) \emph{Random Samp.}, wherein examples are randomly sampled from a predefined candidate set and combined with the prompt template; and 2) \emph{NN Samp.} denotes that in-context examples are chosen through nearest neighbor search using textual branch-based similarities derived from CLIP~\cite{RadfordKHRGASAM21}.
The experimental results on layout generation are illustrated in Figure~\ref{fig:abl_rl}. 
Compared to random sampling, NN sampling generates more accurate layouts, indicating that semantic similarities are helpful for the layout planning of LLMs.
However, closeness in semantics does not mean all of that in layout, \ie, the abilities of language understanding and layout planning may be not the same, and thus different internal mechanisms of LLMs may be triggered.
In contrast, the proposed feedback-based sampling strategy achieves the best performance regarding mIoU and LaySim metrics, demonstrating its effectiveness.

\mysubsubsec{Impact of Shot Number.}
To investigate the impact of the number of in-context examples in activating the layout planning of LLMs, we conduct experiments under zero-shot and few-shot settings (2, 3, 4, and 5).
As seen in Figure~\ref{fig:sensitive_shot}, we first observe that the performance of layout planning exhibits improvement with the increase of the shot number from 0 to 4, signifying that a larger number of in-context examples provide more informative clues, thereby enhancing the performance of LLMs.
When reaching the 3-shot, the performance improvement seems to be saturated, after which LLMs improve slightly. 
Significantly, even under the zero-shot setting, LLMs demonstrate competitive performance, outperforming recent baseline models as indicated in Table~\ref{table:overall_performance}, underscoring the generalization capability of LLMs.

\mysubsubsec{Impact of Relation-aware Image Generation. }
In Section~\ref{sec:Layout-guided-T2I}, we introduce the interaction-based relation-aware image generation module to enhance generation quality.
To delve into how the generation can be affected by this module, we conduct the ablation study by removing it from the full framework, \ie, the original GLIGEN framework, as shown in Figure~\ref{fig:abl_interaction}.
The experimental results on five categories of the re-organized test set on COCO 2014 manifest that the cross-modal interaction among local relation-aware concepts contributes substantively to the relation modeling for text-to-image diffusion models guided by layout information. 
Particularly, considerable performance improvements are observed within ``Semantic'' and ``Mixed'' relation categories, which may be attributable to the high requirement for cross-modal semantic understanding and modeling.

\begin{figure*}[!t]
	\includegraphics[width=0.97\textwidth]{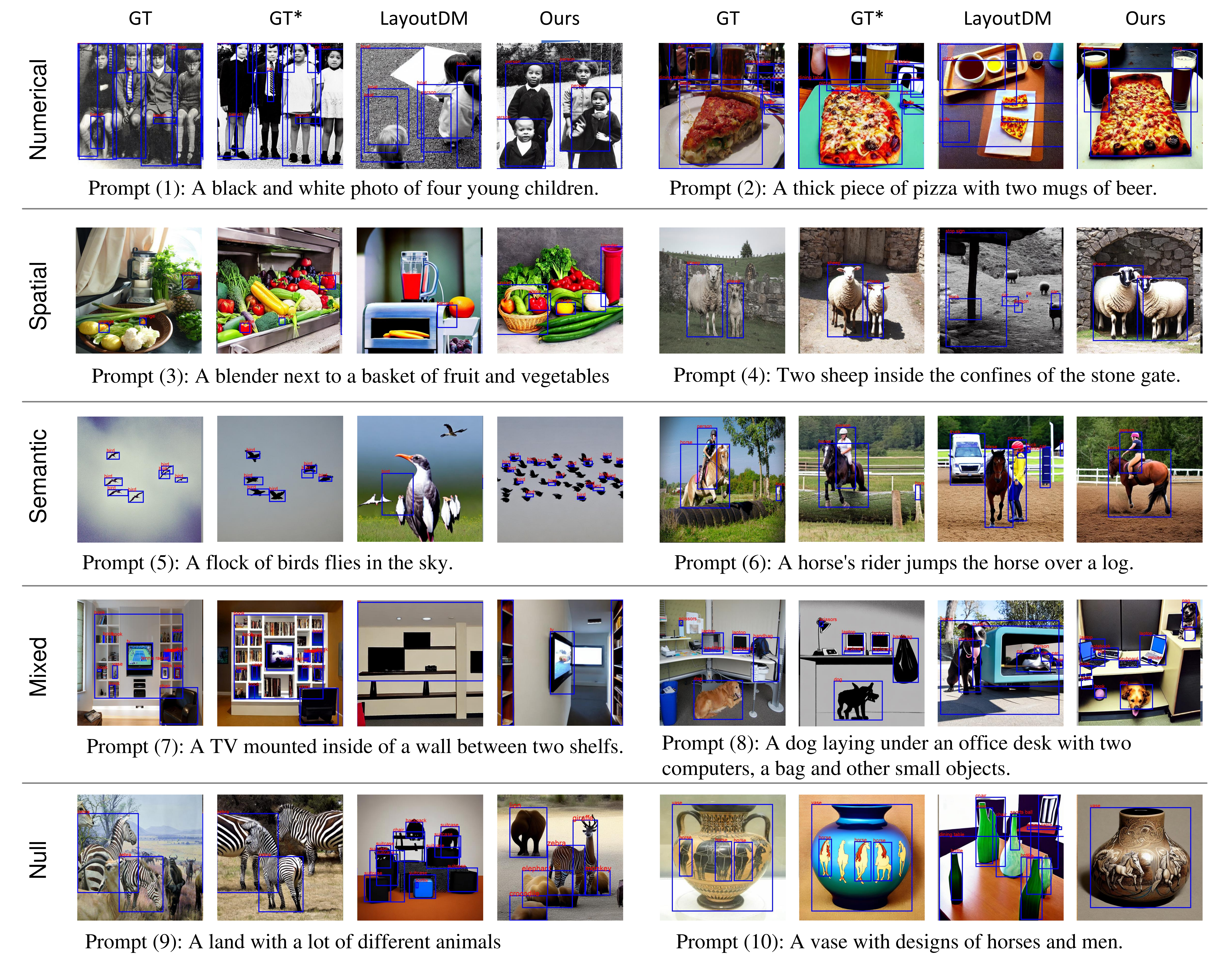}
	\vspace{-3mm}
	\caption{Qualitative results on five test subsets of COCO 2014. Numerical, Spatial, Semantic, Complex, and Abstract layouts are shown from top to bottom. The ground-truth (GT), the ground-truth layout with the generated image (GT*), the result generated from LayoutDM, and our result for each prompt are shown from left to right.}
\label{fig:case}
\vspace{-1.5mm}
\end{figure*}

\subsubsection{\bf Case Study}
To gain an intuitive efficacy of the proposed method, we display some cases from 5 test subsets, as shown in Figure~\ref{fig:case}. By comparing different methods with the ground truth samples. 
We have the following discussions: 
1) A layout of an image plays a key role in the generation process since the prior layout determines logic and overall semantics for the target image.
2) The layout-to-image generation has achieved impressive performance since the generated image given the ground-truth layout is comparable to the real image except for some details.
3) Although the recently proposed LayoutDM~\cite{abs-2303-08137} achieves promising performance on user interface and research paper layout design, it fails to generate satisfying layouts in real-world scenes. 
4) Our proposed method is able to legitimately reason the distribution of objects and precisely depict their relations in the generated images, which demonstrates the effective elicitation of the layout planning capabilities from LLMs.

\section{Conclusion}
In this work, we aim to explore the cross-modal text-guided image generation problem. 
We find existing generative models are weak in layout planning, and propose to tackle this issue from five aspects, including numerical reasoning, spatial relation modeling, semantic relation understanding, complex layout planning, and abstract imagination. Inspired by the recent remarkable success of LLMs, we probe the above abilities via prompting and then further motivate LLMs to achieve layout planning. Concretely, we propose a feedback-based learning strategy to perform in-context learning for LLMs and a relation-aware interaction module to promote image generation. Extensive experiments on the constructed test set validate the effectiveness and superiority of the proposed model.

\bibliographystyle{ACM-Reference-Format}
\balance
\bibliography{acmmm}

 \newpage
\appendix

\twocolumn[
\begin{@twocolumnfalse}
\centering{\huge{\textbf{Appendix}}
\\[10pt]  }
\end{@twocolumnfalse}
]

\section{Extended Technical Details}

\subsection{Fourior Mapping}
Each bounding box is represented as $b = [\alpha_{min}, \beta_{min}, \alpha_{max}, \beta_{max}]$ with its top-left and bottom-right coordinate quadruple.
Recent work \cite{RahamanBADLHBC19} shows that deep networks are biased toward learning lower frequency functions, resulting in performing poorly at representing high-frequency variation in coordinates.
Thus, following \cite{MildenhallSTBRN20,TancikSMFRSRBN20}, we encode bounding box coordinates with the Fourier embedding before feeding them to the network :
\begin{equation}
\begin{aligned}
        \gamma(\rho) = &(\text{sin}(2^0\pi\rho), \text{cos}(2^0\pi\rho), \text{sin}(2^1\pi\rho), \text{cos}(2^1\pi\rho), \\
        &\cdots, \text{sin}(2^{L-1}\pi\rho), \text{cos}(2^{L-1}\pi\rho)) \,,
\end{aligned}
\end{equation}
where function $\gamma(\cdot)$ is applied separately to each of the four coordinate values, and we set $L=32$.

\subsection{The Analysis of Text Prompt}
When synthesizing images conditioned on text prompts, the pivotal thing is that the model should have a comprehensive understanding of the latent intention behind the text prompt, which involves identifying the objects to be generated, their properties, and the relationships between them.
Based on observations, we divide the content in the text prompt into the following components:
\begin{compactitem}
    \item \textbf{Objects}: The specific entities or elements that need to be present in the image, such as human, animal, plant, transportation, building, etc.
    \item \textbf{Attributes}: The specific properties or characteristics of the objects that need to be accurately represented in the image, such as  color, size, shape, texture, or quantity.
    \item \textbf{Relationships}: These describe the connections or interactions between the objects, such as \emph{spatial relationships} (\eg, next to, above, below, left, inside, or contain), \emph{semantic relationships} (\eg, belonging to, interacting with), or \emph{action-based relationships} (\eg, holding, pushing, driving, sitting, lying, or driving).
    \item \textbf{Scene Context}: This refers to the overall context or environment of the scene, including background elements, lighting, style, and other contextual factors.
\end{compactitem}

\begin{table*}[!t]
 \setlength{\tabcolsep}{3mm}
\caption{
    The statistics of the constructed test dataset. `\#Num' denotes the number of the data samples, `\#Avg.bbox' denotes the average number of bounding boxes in an image, `\#Avg.Cap.Len' denotes the average length of the captions in the dataset.
    }
    \scalebox{0.90}{
        \vspace{-2mm}
            \centering
            \renewcommand\arraystretch{1.3}
            \begin{tabular}{l| ccc | l}
            \Xhline{4\arrayrulewidth}
             & \#Num & \#Avg.bbox & \# Avg.Cap.Len  & Caption Examples\\ 
            \hline\hline
            \multirow{2}{*}{Numerical}  & \multirow{2}{*}{155}  & \multirow{2}{*}{6.23} & \multirow{2}{*}{9.55}  & $\bullet$ two old cell phones and a wooden table. \\
            & & & & $\bullet$ two plates some food and a fork knife and spoon. \\
            \hdashline
            \multirow{2}{*}{Spatial} &   \multirow{2}{*}{200}   & \multirow{2}{*}{5.35} & \multirow{2}{*}{10.25} & $\bullet$ a large clock tower next to a small white church.  \\
            & & & & $\bullet$ a bowl with some noodles inside of it. \\
            \hdashline
            \multirow{2}{*}{Semantic}  & \multirow{2}{*}{200} & \multirow{2}{*}{7.10} & \multirow{2}{*}{10.62}  & $\bullet$ a train on a track traveling through a countryside.   \\
            & & & & $\bullet$ a living room filled with couches, chairs, tv, and windows. \\
            \hdashline
            \multirow{2}{*}{Mixed} & \multirow{2}{*}{188}  & \multirow{2}{*}{6.94} & \multirow{2}{*}{10.76} & $\bullet$ one motorcycle rider riding going up the mountain two going down. \\
            & & & & $\bullet$ a group of three bathtubs sitting next to each other. \\
            \hdashline
            \multirow{2}{*}{Null} & \multirow{2}{*}{200}  & \multirow{2}{*}{6.17} & \multirow{2}{*}{9.62} &  $\bullet$ a kitchen scene complete with a dishwasher, sink and an oven.\\
            & & & & $\bullet$ a person with a hat and some ski poles. \\
            \hdashline
            Total & 943  & 6.35 & 10.18 & - \\
            \bottomrule
            \end{tabular}
    }
\label{table:dataset}
\end{table*}

\vspace{4mm}
\section{Experiment Settings}

\subsection{Detailed Implementation Settings}
\label{appendix:detailed settings}
We adopt a two-stage strategy to optimize the proposed framework. 
In the first stage,  we use a Scene Graph parser (\url{https://github.com/vacancy/SceneGraphParser}) to extract the `\emph{subject-predicate-object}' triplets for each caption, the maximum number of triplets is 10, and then obtain their embeddings using CLIP textual branch with the ``clip-vit-large-patch14'' version. Take as input the triplet embeddings and intermediate representations in the UNet of Latent Diffusion model, multi-head cross-attention layers are plugged to perform relation-aware interaction. Then, we perform continual learning based on the pre-trained GLIGEN model, with an initial learning of 3e-5, and a batch size of 1. 

In the second stage, the key is to learn an optimized policy $\pi_\psi (c_i | y_i)$ to select informative in-context examples. Concretely, we implement $f (\cdot)$ in Eq.(\ref{eqn:policy_implemented}) with a linear layer with 128 hidden neurons which is optimized to learn layout-level similarities on the top of semantic embeddings induced from the CLIP textual branch. 
we optimize the feedback-based sampler via Reinforcement Learning. Two-fold feedback is considered for policy gradient, including layout-level reward $R^B$ (implemented with mIoU) and image-level reward $R^I$ (consisting of image-to-text similarities, image-to-image similarities, and aesthetic scores). Considering the different numerical scales and distributions, we apply balancing factors to reweight each term: $R = 10 \cdot \text{mIoU} + \text{Sim} + 0.1 \cdot \text{Aes}$. 
We use the \textit{gpt-3.5-turbo} model via OpenAI API (\url{https://platform.openai.com/docs/models/gpt-3-5}), considering its powerful language understanding and reasoning abilities.  Under the few-shot setting, we randomly sample 64 instances for training, and 32 instances to form the candidate set. Besides, we set 2 as the shot number by default. During the optimization phase, the total number of epochs, the batch size, and the initial learning rate are set to 80, 8, and $2 \times 10^{-4}$, respectively.


As for baseline methods, considering some layout generation models are unconditional or other types of conditions (\eg, partial labels and simple phrases) instead of a complete free-form natural language, we add extra cross-attention layers and train them on the COCO 2014 dataset again.


\vspace{4mm}
\subsection{Detailed Test Set Construction}
\label{appendix:testset construction}

To thoroughly assess the layout planning and relation understanding abilities, we construct a new test set from the raw COCO 2014 validation set $U$.
Concretely, we build the test set in four steps:

\mysubsubsec{1. Pre-define Filtering Rules.} Specifically, 
    we choose data samples whose captions include specific keywords to construct \texttt{numeral}, \texttt{spatial} subset and use the NLP toolkit spacy (\url{https://spacy.io/models/en\#en\_core\_web\_sm}) to parse captions and build \texttt{semanitic} subset according to Part-of-Speech (POS) tagging:
    \begin{compactitem}
        \item[a.] The keywords list for filtering the captions containing the numeral is: \emph{"two", "three", "four", "five", "six", "seven", "eight", "nine", "ten", "many", "bunch", "some", "several", "various", "group"}.
        \item[b.] The keywords list for  filtering the captions containing the spatial relationship is: \emph{"left", "right", "top", "down", "near", "next", "side", "above", "inside", "outside", "below", "front", "back", "under", "around", "bottom", "up", "beside", "beneath", "underneath"}.      
        \item[c.] Generally, a caption including notional verbs is highly possible to depict some semantic relations. To decide whether a caption contains any notional verbs, we first find words with ``VERB'' POS. If a word is neither detected as an auxiliary or model verb using dependency labels nor detected as a linking verb, then it is viewed as a notional verb. 
    \end{compactitem}
\mysubsubsec{2. Primary Screening.} We screen the valid dataset according to the keywords, constructing three primary screening datasets, \ie, the numerical dataset ($\widehat{U}_{num}$), the spatial dataset ($\widehat{U}_{spa}$), and the semantic dataset ($\widehat{U}_{sem}$).

\mysubsubsec{3. Second Filtering.} 
    \begin{compactitem}
        \item[a.] To construct the \texttt{only numerical} subset that only contains the numeral in the captions, we exclude the instances from the numerical dataset that also appears in the spatial and semantic datasets, \ie, $U_{num} = \widehat{U}_{num} - \widehat{U}_{spa} - \widehat{U}_{sem}$. 
        Similarly, we build the \texttt{only semantic} dataset (\ie, $U_{sem} = \widehat{U}_{sem} - \widehat{U}_{spa} - \widehat{U}_{num}$) and the \texttt{only spatial} dataset (\ie, $U_{spa} = \widehat{U}_{spa} - \widehat{U}_{sem} - \widehat{U}_{num}$).
        \item[b.] We take the intersection of numeral, spatial, and semantic relationship datasets as the \texttt{Mixed} dataset, \ie, $U_{mix} = \widehat{U}_{spa} \land \widehat{U}_{sem} \land  \widehat{U}_{num}$. 
        \item[c.] To construct the \texttt{Null} datasets that do not contain any explicit relation keywords in prompts, we filter the instances included in the numeral, spatial, and semantic dataset from the total dataset, \ie, $U_{Null} = U - \widehat{U}_{num} - \widehat{U}_{spa} - \widehat{U}_{sem}$.
    \end{compactitem}
\mysubsubsec{4. Sampling.} For a dataset with more than 200 instances, we randomly select a subset of 200 instances as the final dataset.
Finally, the statistics of the constructed test dataset are shown in Table \ref{table:dataset}.

\subsection{Detailed Layout and Image Evaluation}
\label{apendix:detailed evaluation}
For quantitative experiments, we consider various metrics from different aspects to evaluate our method on layout generation and image generation.
We now introduce these metrics as follows.

\mysubsubsec{Layout Evaluation:}
\begin{compactitem}
        \item \textbf{Fr\'echet Inception Distance} (FID)~\cite{HeuselRUNH17}. Following \cite{abs-2303-08137}, we first train a Transformer-based model that can extract discriminative layout features, which is then utilized to compute the FID.
   \item \textbf{Maximum IoU (mIoU)}. This score evaluates the overlap of the ground layout and predicted layout.  
   \item \textbf{Layout Similarity (LaySim)}. LaySim proposed by \cite{PatilBPA20} aims to measure the similarity between the generated layout and the given layout. Specifically, given the generated layout $B$ and gold layout $B^{'}$, we first assign a weighted edge between any pair of bounding boxes $b \in B$ and $b^{'} \in B^{'}$, indicating how similar $b$ and $b^{'}$ are in terms of shape. Then, we calculate as the final score the aggregated weight of the maximum (weighted) matching between the layouts $B$ and $B^{'}$.   
    \end{compactitem}
    
Note that mIoU and LaySim are calculated based on close-set labels, while our method generates free labels for each bounding box. To obtain mIoU and LaySim of our method, we employ CLIP textual branch to compute semantic similarities between the predicted labels and the pre-defined 80 classes, and then map each free label to the closest pre-defined one.

\mysubsubsec{Image Evaluation:}
\begin{compactitem}
       \item \textbf{Fr\'echet Inception Distance} (FID)~\cite{HeuselRUNH17}. FID measures the Fr\'echet Distance between the distribution of the synthetic images and real-world images in the feature space of a pre-trained Inception v3 network. Practically, we employ \emph{torch-fidelity} (\url{https://github.com/mseitzer/pytorch-fid}) to calculate the FID score.  
       \item \textbf{Cross-modal (Sim(I-T)) and Intra-model Similarity (Sim(I-I))}. In order to perform the quantitative analysis on the semantic correctness of the text-to-image generation, we take into account the similarity scores via BLIP-2 \cite{li2023blip}. Concretely, we calculate the two types of semantic similarities between generated images and given text prompt, and gold images, respectively. 
\end{compactitem}

\subsection{Detailed Prompts}
Here, we show the detailed prompt examples used during in-context learning under 0/1/2/3-shot settings:

\mybullet{\textbf{0-shot prompt.}}
\vspace{-2mm}

\begin{mybox}
\texttt
\textbf{\texttt{[Instruction]}}. Now you are an assistant to help me design a layout given a description. Concretely, a layout denotes a set of "object: bounding box" items. "object" means any object name in the world, while "bounding box" is formulated as [x, y, w, h], where "x, y" denotes the top left coordinate of the bounding box, "w" denotes the width, and "h" denotes the height. The six values "x, y, w, h, x+w, y+h" are all larger than 0 and smaller than 1. Next, I will give you an input that describes an image, and then you should give me an output with the format "\\
output:\\
object: [x, y, w, h], \\
object: [x, y, w, h],\\
...\\
".
\textbf{\texttt{[Test]}}. \\
input: An open refrigerator with food and condiments inside of it.
\begin{mybluebox}
\texttt
\textbf{\texttt{[Layout]}}. \\
output: \\
food: [0.1, 0.2, 0.4, 0.3], \\
food: [0.6, 0.2, 0.3, 0.4], \\
condiments: [0.1, 0.6, 0.4, 0.2], \\
condiments: [0.6, 0.6, 0.3, 0.3], \\
refrigerator: [0, 0, 1, 1]
\end{mybluebox}
\end{mybox}

\begin{mybox}
\texttt
\textbf{\texttt{[Instruction]}}. Now you are an assistant to help me design a layout given a description. Concretely, a layout denotes a set of "object: bounding box" items. "object" means any object name in the world, while "bounding box" is formulated as [x, y, w, h], where "x, y" denotes the top left coordinate of the bounding box, "w" denotes the width, and "h" denotes the height. The six values "x, y, w, h, x+w, y+h" are all larger than 0 and smaller than 1. Next, I will give you an input that describes an image, and then you should give me an output with the format "\\
output:\\
object: [x, y, w, h], \\
object: [x, y, w, h],\\
...\\
".
\textbf{\texttt{[Test]}}. \\
input: a person with a hat and some ski poles.
\begin{mybluebox}
\texttt
\textbf{\texttt{[Layout]}}. \\
output: \\
person: [0.2, 0.1, 0.4, 0.8], \\
hat: [0.3, 0.05, 0.3, 0.2], \\
ski pole 1: [0.6, 0.3, 0.05, 0.6], \\
ski pole 2: [0.35, 0.3, 0.05, 0.6]
\end{mybluebox}
\end{mybox}

\mybullet{\textbf{1-shot prompt.}}
\vspace{-2mm}
\begin{mybox}
\texttt
\textbf{\texttt{[Instruction]}}. Now you are an assistant to help me design a layout given a description. Concretely, a layout denotes a set of "object: bounding box" items. "object" means any object name in the world, while "bounding box" is formulated as [x, y, w, h], where "x, y" denotes the top left coordinate of the bounding box, "w" denotes the width, and "h" denotes the height. The six values "x, y, w, h, x+w, y+h" are all larger than 0 and smaller than 1. Next, I will give you several examples for you to understand this task.\\
\textbf{\texttt{[In-context Examples]}}. \\
input: a kitchen with low lights and allot on the counters.\\
output: \\
knife: [0.22, 0.48, 0.02, 0.02] \\
knife: [0.2, 0.45, 0.02, 0.02] \\
knife: [0.22, 0.45, 0.02, 0.03] \\
knife: [0.21, 0.47, 0.02, 0.02] \\
sink: [0.34, 0.51, 0.42, 0.05] \\
knife: [0.19, 0.45, 0.03, 0.03] \\
spoon: [0.03, 0.47, 0.04, 0.04] \\
oven: [0.01, 0.61, 0.25, 0.39] \\
knife: [0.17, 0.45, 0.04, 0.03] \\ 
knife: [0.21, 0.48, 0.02, 0.03] \\
knife: [0.2, 0.49, 0.02, 0.02] \\
knife: [0.19, 0.48, 0.02, 0.02] \\
\\
\textbf{\texttt{[Test]}}. \\
input: An open refrigerator with food and condiments inside of it.
\begin{mybluebox}
    \texttt
    \textbf{\texttt{[Layout]}}. \\
output:\\
refrigerator: [0.1, 0.1, 0.4, 0.8] \\
milk: [0.15, 0.2, 0.1, 0.1] \\
eggs: [0.25, 0.3, 0.1, 0.1] \\
cheese: [0.35, 0.2, 0.1, 0.1] \\
mayonnaise: [0.15, 0.5, 0.1, 0.1] \\
ketchup: [0.25, 0.6, 0.1, 0.1] \\
lettuce: [0.35, 0.5, 0.1, 0.1]
\end{mybluebox}

\end{mybox}

\begin{mybox}
\texttt
\textbf{\texttt{[Instruction]}}. Now you are an assistant to help me design a layout given a description. Concretely, a layout denotes a set of "object: bounding box" items. "object" means any object name in the world, while "bounding box" is formulated as [x, y, w, h], where "x, y" denotes the top left coordinate of the bounding box, "w" denotes the width, and "h" denotes the height. The six values "x, y, w, h, x+w, y+h" are all larger than 0 and smaller than 1. Next, I will give you several examples for you to understand this task.\\
\textbf{\texttt{[In-context Examples]}}. \\
input: A yield sign followed by a stop sign on a deserted road. \\
output: \\
\indent\indent stop sign: [0.23, 0.48, 0.05, 0.07] \\
\\
\textbf{\texttt{[Test]}}. \\
input: a person with a hat and some ski poles.
\begin{mybluebox}
    \texttt
    \textbf{\texttt{[Layout]}}. \\
output: \\
person: [0.4, 0.1, 0.2, 0.8]  \\
hat: [0.45, 0.05, 0.1, 0.1] \\
ski pole 1: [0.35, 0.6, 0.05, 0.3]  \\
ski pole 2: [0.55, 0.6, 0.05, 0.3]
\end{mybluebox}
\end{mybox}

\mybullet{\textbf{2-shot prompt.}}

\begin{mybox}
\texttt
\textbf{\texttt{[Instruction]}}. Now you are an assistant to help me design a layout given a description. Concretely, a layout denotes a set of "object: bounding box" items. "object" means any object name in the world, while "bounding box" is formulated as [x, y, w, h], where "x, y" denotes the top left coordinate of the bounding box, "w" denotes the width, and "h" denotes the height. The six values "x, y, w, h, x+w, y+h" are all larger than 0 and smaller than 1. Next, I will give you several examples for you to understand this task.\\
\textbf{\texttt{[In-context Examples]}}. \\
input: Guy walking an elephant down a dirt path. \\
output: \\
person: [0.37, 0.43, 0.19, 0.56] \\
elephant: [0.47, 0.0, 0.41, 0.98] \\
\\
input: a black cow looking over an iron fence. \\
output:  \\
cow: [0.09, 0.23, 0.77, 0.66] \\
cow: [0.74, 0.7, 0.24, 0.14] \\
\\
\textbf{\texttt{[Test]}}. \\
input: Three zebra and four giraffes inside a fenced area.
\begin{mybluebox}
    \texttt
    \textbf{\texttt{[Layout]}}. \\
output: \\
zebra1: [0.1, 0.2, 0.2, 0.6] \\
zebra2: [0.3, 0.3, 0.2, 0.6] \\
zebra3: [0.5, 0.2, 0.2, 0.6] \\
giraffe1: [0.1, 0.8, 0.3, 0.2] \\
giraffe2: [0.4, 0.8, 0.3, 0.2]  \\
giraffe3: [0.7, 0.8, 0.3, 0.2]  \\
giraffe4: [0.4, 0.6, 0.3, 0.2]  \\
fence: [0.0, 0.0, 1.0, 1.0]
\end{mybluebox}

\end{mybox}

\begin{mybox}
\texttt
\textbf{\texttt{[Instruction]}}. Now you are an assistant to help me design a layout given a description. Concretely, a layout denotes a set of "object: bounding box" items. "object" means any object name in the world, while "bounding box" is formulated as [x, y, w, h], where "x, y" denotes the top left coordinate of the bounding box, "w" denotes the width, and "h" denotes the height. The six values "x, y, w, h, x+w, y+h" are all larger than 0 and smaller than 1. Next, I will give you several examples for you to understand this task.\\
\textbf{\texttt{[In-context Examples]}}. \\
input: Guy walking an elephant down a dirt path. \\
output: \\
person: [0.37, 0.43, 0.19, 0.56] \\
elephant: [0.47, 0.0, 0.41, 0.98]  \\
 \\
input: A man holding a tennis racquet on top of a tennis court.\\
output: \\
person: [0.5, 0.09, 0.45, 0.89] \\
tennis racket: [0.65, 0.1, 0.17, 0.12] \\
\\
\textbf{\texttt{[Test]}}. \\
input: a person with a hat and some ski poles.
\begin{mybluebox}
    \texttt
    \textbf{\texttt{[Layout]}}. \\
output: \\
person: [0.35, 0.4, 0.3, 0.6] \\
hat: [0.4, 0.35, 0.2, 0.15] \\
ski pole 1: [0.6, 0.4, 0.05, 0.5] \\
ski pole 2: [0.35, 0.4, 0.05, 0.5] 
\end{mybluebox}
\end{mybox}

\vspace{-10mm}
\mybullet{\textbf{3-shot prompt.}}

\begin{mybox}
\texttt
\textbf{\texttt{[Instruction]}}. Now you are an assistant to help me design a layout given a description. Concretely, a layout denotes a set of "object: bounding box" items. "object" means any object name in the world, while "bounding box" is formulated as [x, y, w, h], where "x, y" denotes the top left coordinate of the bounding box, "w" denotes the width, and "h" denotes the height. The six values "x, y, w, h, x+w, y+h" are all larger than 0 and smaller than 1. Next, I will give you several examples for you to understand this task.\\
\textbf{\texttt{[In-context Examples]}}. \\
input: A notebook, mp3 player, pencil, pen, wallet, purse, and a cell phone. \\
output:  \\
bed: [-0.0, 0.01, 0.99, 0.97]\\
cell phone: [0.64, 0.07, 0.15, 0.16]\\
handbag: [0.36, 0.04, 0.25, 0.2]\\
handbag: [0.05, 0.02, 0.3, 0.25]\\
book: [0.0, 0.21, 0.48, 0.71]\\
handbag: [0.8, 0.08, 0.2, 0.3]\\
\\
input: A kitchen scene with a lot of items on the counters.\\
output: \\
potted plant: [0.3, 0.31, 0.09, 0.15]\\
oven: [0.62, 0.33, 0.3, 0.54]\\
sink: [0.35, 0.45, 0.17, 0.04]\\
cup: [0.75, 0.3, 0.02, 0.04]\\
cup: [0.72, 0.3, 0.03, 0.04]\\
bottle: [0.5, 0.34, 0.02, 0.13]\\
spoon: [0.84, 0.35, 0.03, 0.07]\\
microwave: [0.06, 0.35, 0.16, 0.13]\\
vase: [0.31, 0.43, 0.03, 0.04]\\
\\
input: a kitchen with low lights and allot on the counters\\
output: \\
knife: [0.22, 0.48, 0.02, 0.02]\\
knife: [0.2, 0.45, 0.02, 0.02]\\
knife: [0.22, 0.45, 0.02, 0.03]\\
knife: [0.21, 0.47, 0.02, 0.02]\\
sink: [0.34, 0.51, 0.42, 0.05]\\
knife: [0.19, 0.45, 0.03, 0.03]\\
spoon: [0.03, 0.47, 0.04, 0.04]\\
oven: [0.01, 0.61, 0.25, 0.39]\\
knife: [0.17, 0.45, 0.04, 0.03]\\
knife: [0.21, 0.48, 0.02, 0.03]\\
knife: [0.2, 0.49, 0.02, 0.02]\\
knife: [0.19, 0.48, 0.02, 0.02]\\
\\
\textbf{\texttt{[Test]}}. \\
input: A kitchen with an oven, stove, sink, microwave, and refrigerator. 
\begin{mybluebox}
    \texttt
    \textbf{\texttt{[Layout]}}. \\
output: \\
oven: [0.01, 0.2, 0.3, 0.6] \\
stove: [0.35, 0.4, 0.3, 0.2] \\
sink: [0.6, 0.5, 0.3, 0.1] \\
microwave: [0.7, 0.2, 0.2, 0.2] \\
refrigerator: [0.8, 0.4, 0.2, 0.6]
\end{mybluebox}
\end{mybox}

\begin{mybox}
\texttt
\textbf{\texttt{[Instruction]}}. Now you are an assistant to help me design a layout given a description. Concretely, a layout denotes a set of "object: bounding box" items. "object" means any object name in the world, while "bounding box" is formulated as [x, y, w, h], where "x, y" denotes the top left coordinate of the bounding box, "w" denotes the width, and "h" denotes the height. The six values "x, y, w, h, x+w, y+h" are all larger than 0 and smaller than 1. Next, I will give you several examples for you to understand this task.\\
\textbf{\texttt{[In-context Examples]}}. \\
input: A baseball player swinging a bat on top of field.\\
output: \\
person: [0.42, 0.36, 0.22, 0.47]\\
person: [0.18, 0.52, 0.25, 0.36]\\
person: [0.13, 0.44, 0.04, 0.17]\\
person: [0.2, 0.42, 0.05, 0.16]\\
baseball glove: [0.41, 0.66, 0.05, 0.09]\\
baseball bat: [0.61, 0.46, 0.05, 0.01]\\
person: [0.0, 0.41, 0.12, 0.48]\\
\\
input: Guy walking an elephant down a dirt path.\\
output: \\
person: [0.37, 0.43, 0.19, 0.56]\\
elephant: [0.47, 0.0, 0.41, 0.98]\\
\\
input: A man holding a tennis racquet on top of a tennis court.\\
output: \\
person: [0.5, 0.09, 0.45, 0.89]\\
tennis racket: [0.65, 0.1, 0.17, 0.12]\\
\\
\textbf{\texttt{[Test]}}. \\
input: A group of three giraffe standing inside of a cage. 
\begin{mybluebox}
    \texttt
    \textbf{\texttt{[Layout]}}. \\
output: \\
giraffe: [0.1, 0.1, 0.3, 0.8]  \\
giraffe: [0.4, 0.2, 0.3, 0.7] \\
giraffe: [0.7, 0.3, 0.3, 0.6] \\
cage: [0.05, 0.05, 0.9, 0.9]
\end{mybluebox}
\end{mybox}

\section{Experimental Results}
\begin{figure*}[t]
    \subfigure[mIoU]{
		\label{fig:sample_miou}
		\includegraphics[width=0.80\textwidth]{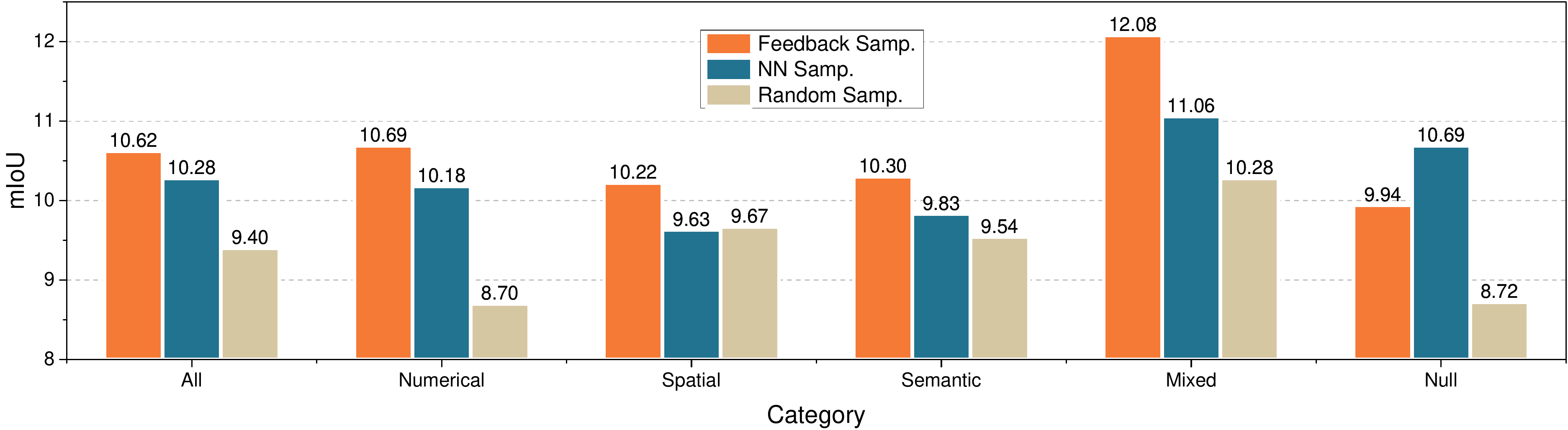}}	
    \subfigure[LaySim]{
		\label{fig:sample_laysim}
		\includegraphics[width=0.80\textwidth]{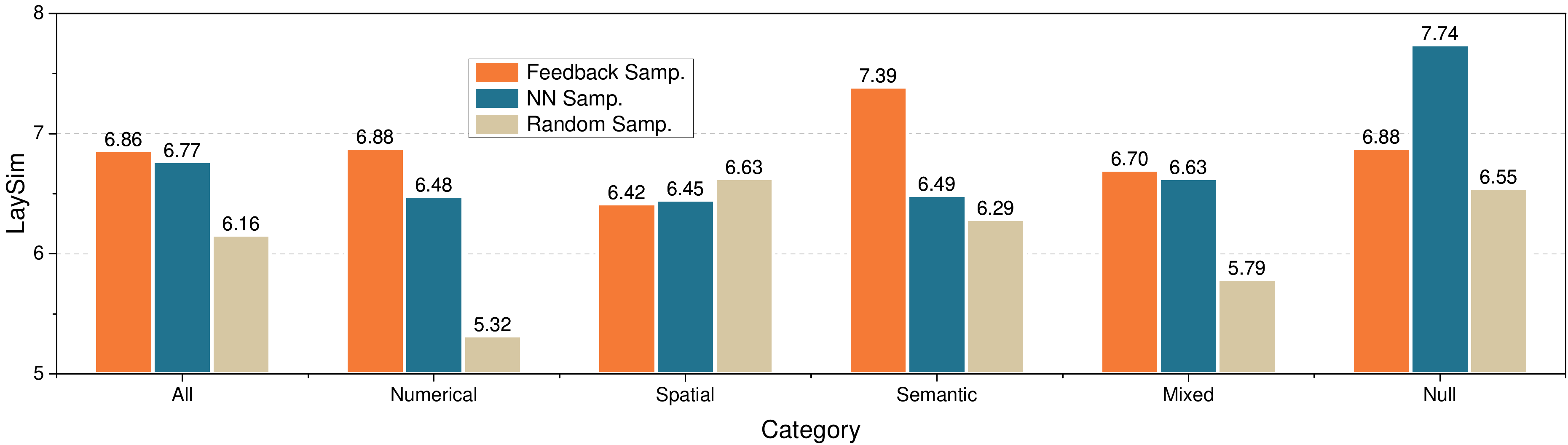}}
    \subfigure[Sim (I-T)]{
		\label{fig:sample_simit}
		\includegraphics[width=0.80\textwidth]{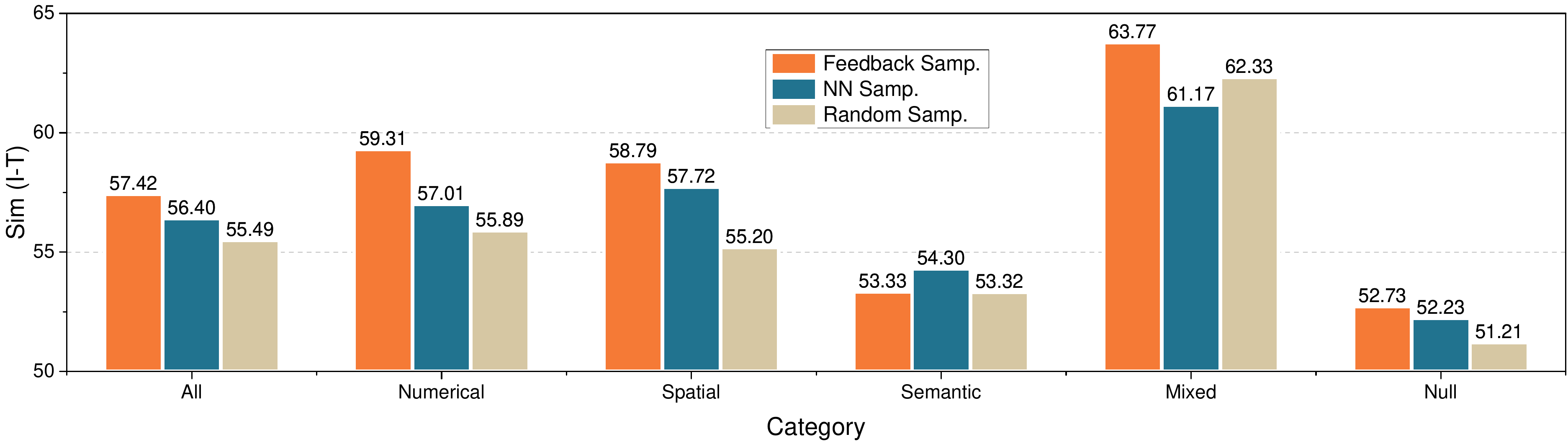}}
	\vspace{-3mm}
	\caption{Ablation study of in-context example sampling on the full test set and five categories. Layout evaluation metrics (a) mIoU$\uparrow$, (b) LaySim$\uparrow$, and image evaluation metric (c) Sim (I-T)$\uparrow$ are reported.  }
\label{fig:abl_sample_miou_laysim}
\vspace{-2mm}
\end{figure*}

\mysubsubsec{Impact of In-context Example Sampling.}
We report the experimental results and performance comparison of Feedback Sampling (Ours), Nearest Neighbor Sampling, and Random Sampling on the full test set and five categories, as shown in Figure~\ref{fig:abl_sample_miou_laysim}. 
Based on these results, we have the following discussions: 
1) In general, the proposed feedback-based sampling performs better than the other two variants in most categories and evaluation metrics. It validates the effectiveness of the proposed sampling strategy. 
2) The layouts generated by Random Sampling are the worst in most cases, especially in the numerical subset. The comparison results show that NN Sampling is able to provide informative in-context examples to some extent and performs better than Random Sampling. Meanwhile, compared with other categories, the numerical subset depends more heavily on the selection of in-context examples. 
3) As for the image evaluation metric Sim (I-T) shown in Figure \ref{fig:sample_simit}, all the three variants in the ``mixed'' relation category perform best, while worst in the ``null'' category. It may be attributable to more contributions of abundant relations in the ``mixed'' category to the textual faithfulness.
And 4) NN Sampling performs best in the Null category according to both mIoU and LaySim metrics. The reason may be that this category does not rely heavily on layout planning abilities and semantic closeness measured by CLIP in NN Sampling is more helpful for the selection of in-context examples.

\begin{figure*}[t]
    \subfigure[FID]{
		\label{fig:shot_fid}
		\includegraphics[width=0.30\textwidth]{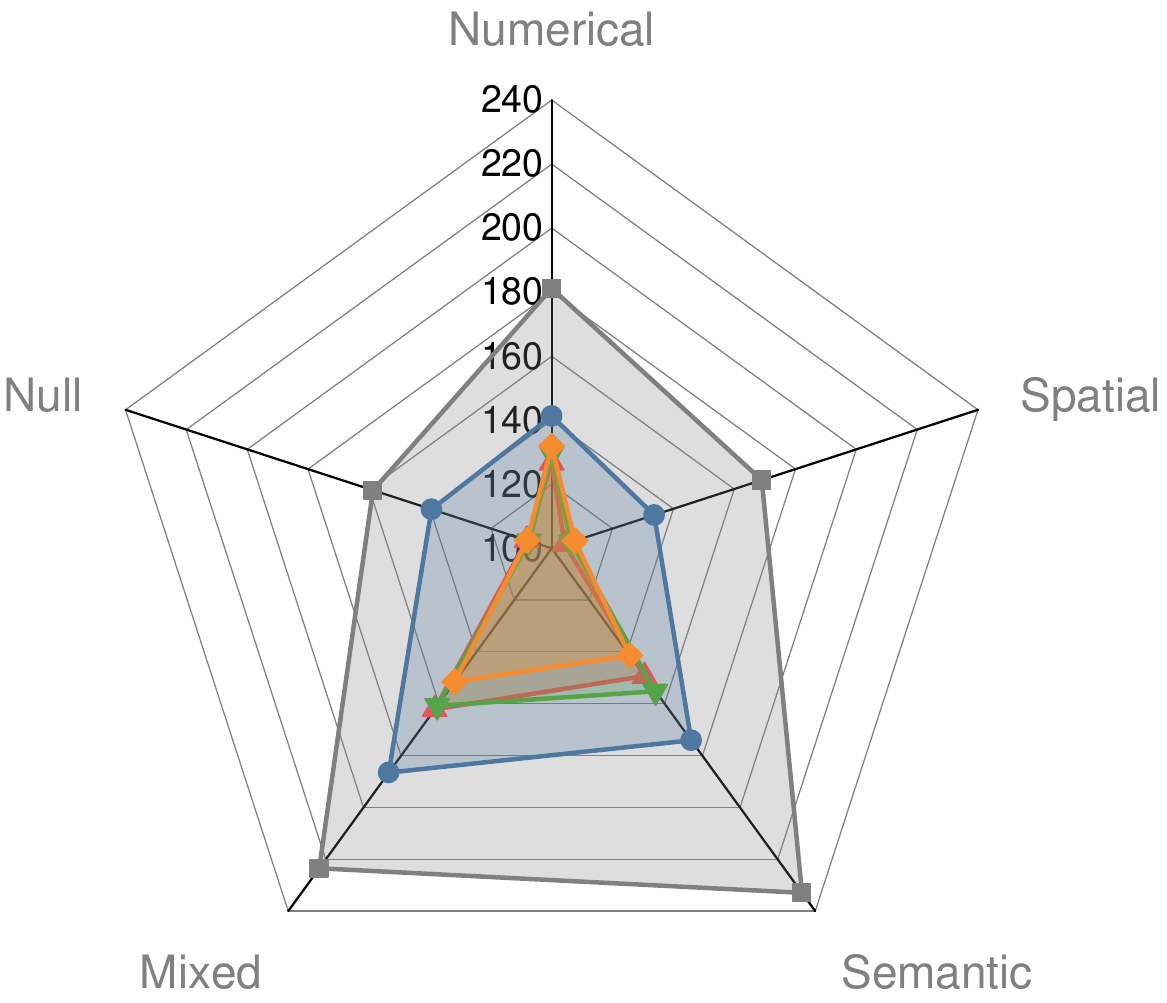}}	
    \subfigure[mIoU]{
		\label{fig:shot_miou}
		\includegraphics[width=0.30\textwidth]{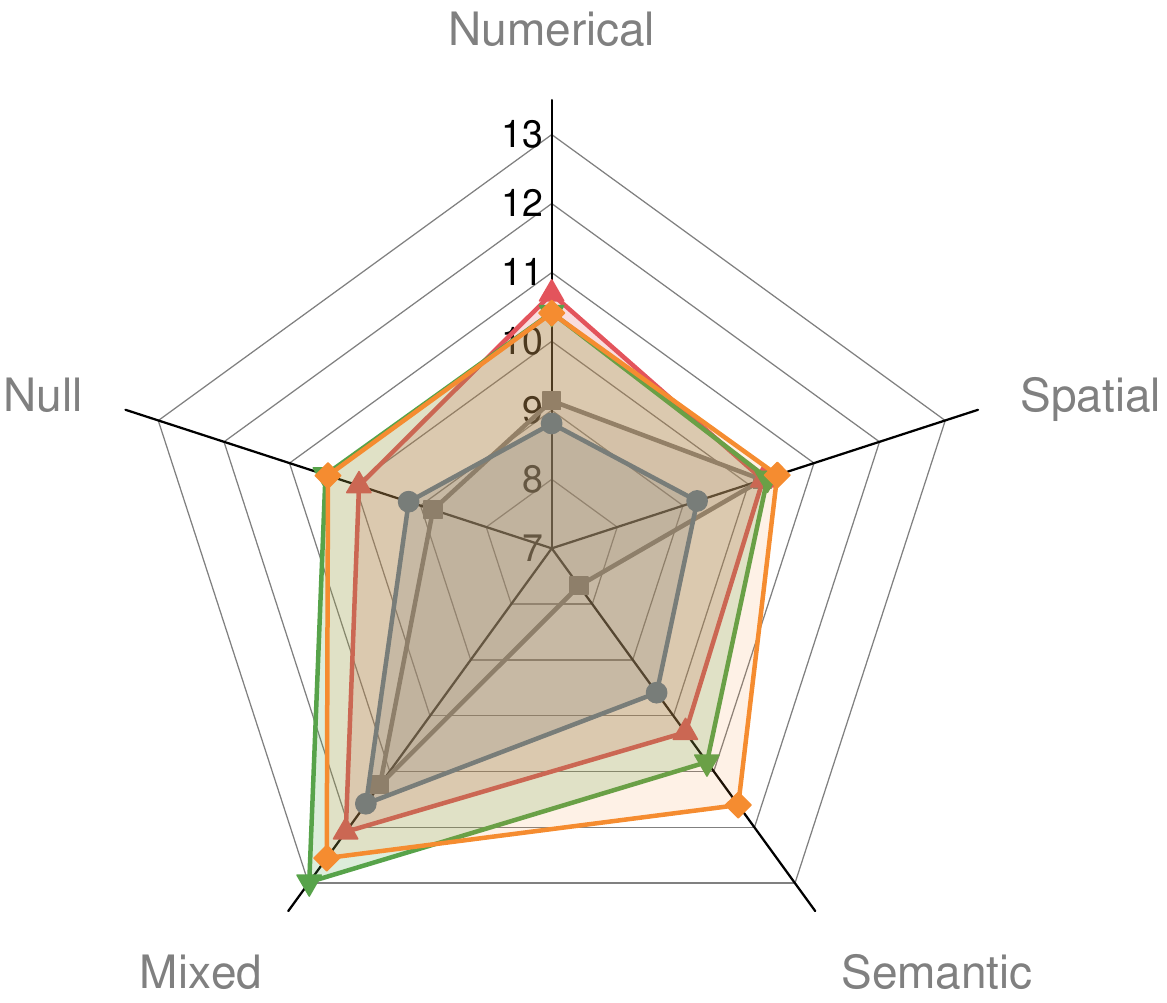}}
    \subfigure[LaySim]{
		\label{fig:shot_laysim}
		\includegraphics[width=0.32\textwidth]{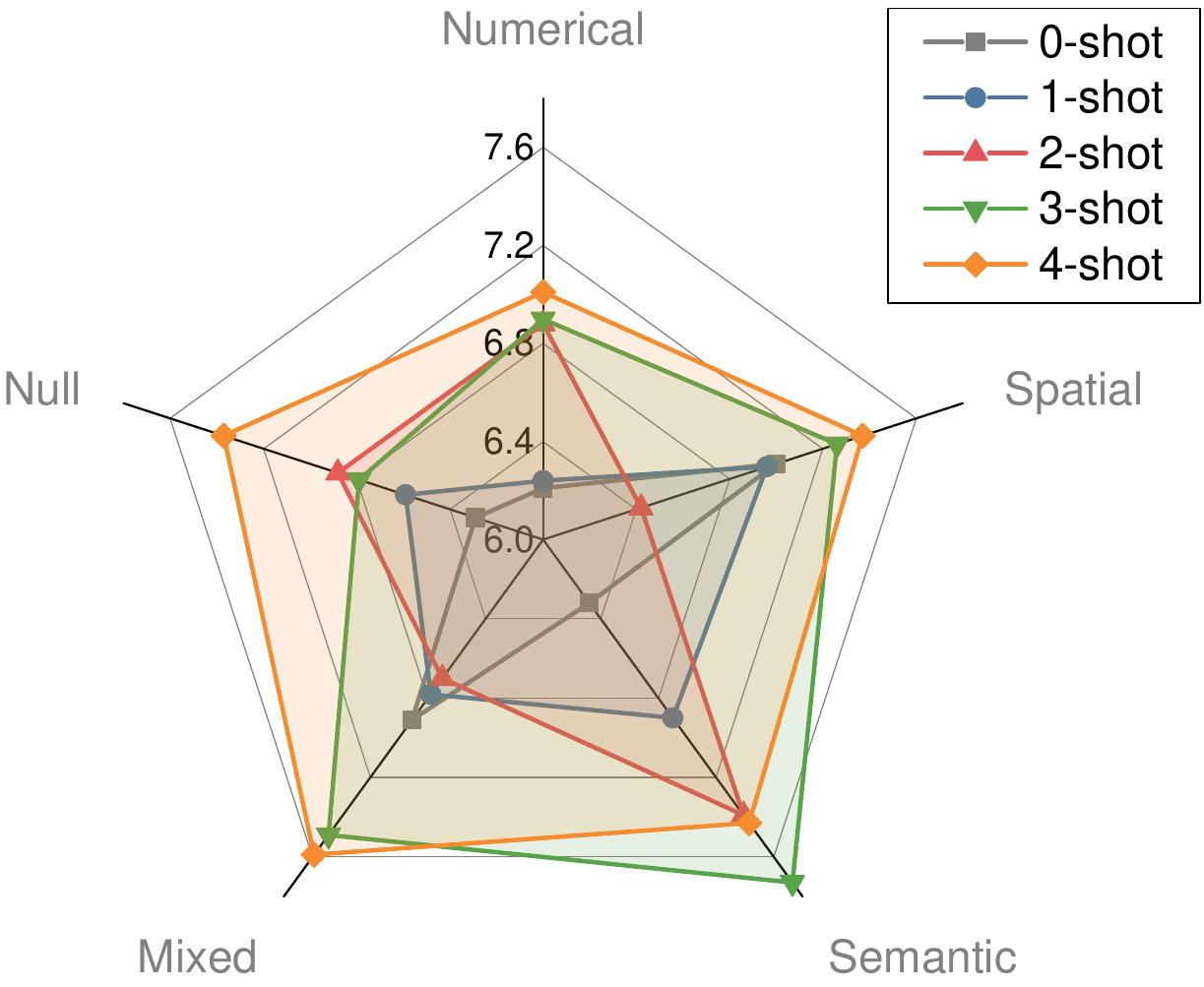}}
	\vspace{-3mm}
	\caption{Layout performance comparison with different shot numbers. Results on three layout evaluation metrics (a) FID$\downarrow$, (b) mIoU$\uparrow$, and (c) LaySim$\uparrow$ are reported. }
\label{fig:ana_shot}
\vspace{-2mm}
\end{figure*}

\mysubsubsec{Impact of Shot Number.}
As shown in Figure~\ref{fig:ana_shot}, we carry out extensive experiments to explore the influence of shot numbers on the layout planning process across five test subsets. All the experiments consistently show that the layout generation performance is sensitive to the shot number, verifying the necessity of using sufficient in-context examples to activate certain abilities of LLMs. Despite this, striving for a balance between the shot number and inference cost should also be considered in practice.

\subsection{More Examples}
\begin{figure*}[t]
	\includegraphics[width=0.80\textwidth]{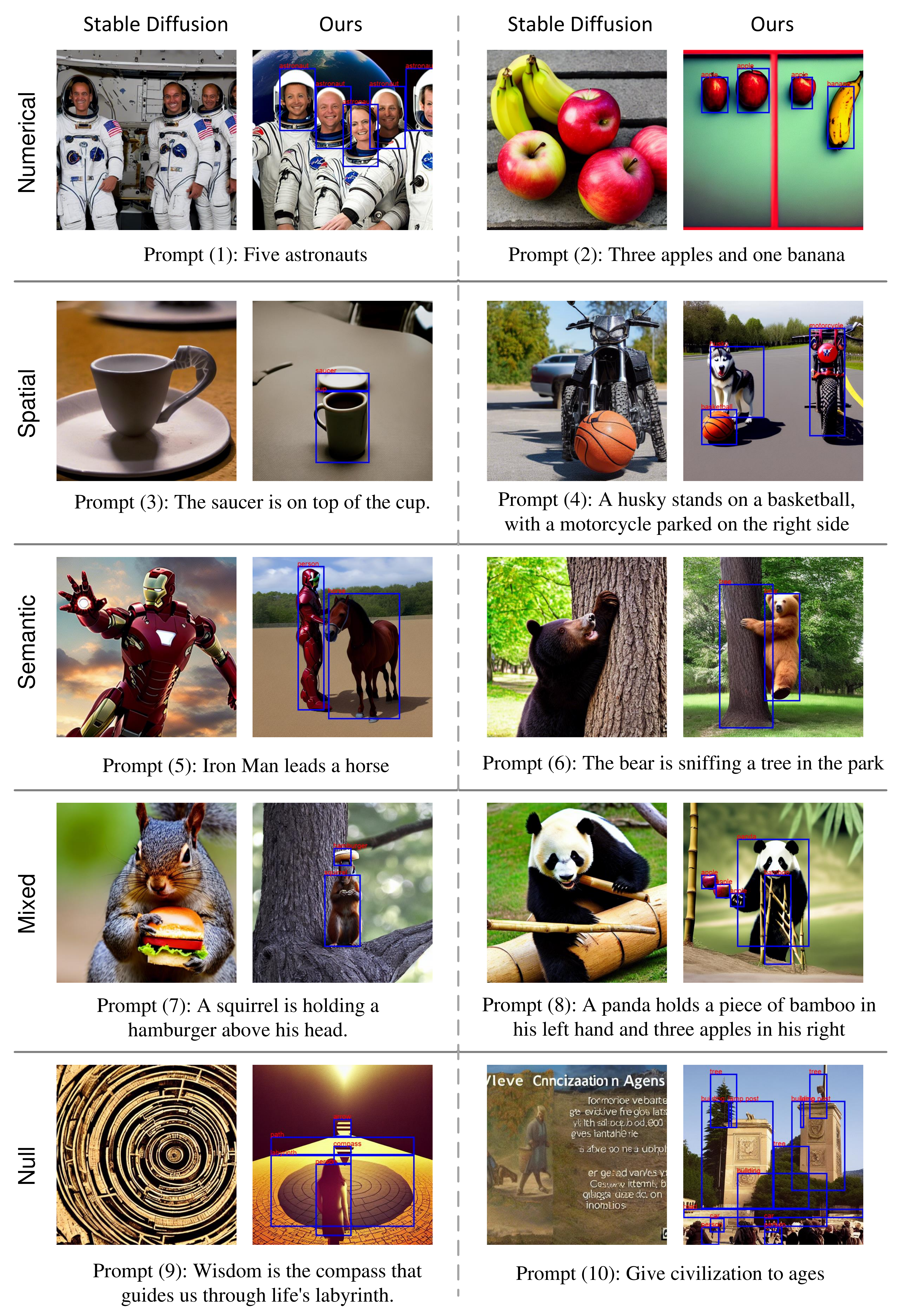}
	\caption{Qualitative comparison between Stable Diffusion and the proposed method on the Numerical, Spatial, Semantic, Mixed, and Null test subset are shown from top to bottom. }
\label{fig:sd_vs_ours}
\vspace{-2mm}
\end{figure*}

\mysubsubsec{Stable Diffusion vs. Ours.}
To compare Stable Diffusion~\cite{RombachBLEO22} and our method in terms of textual faithfulness, we design ten representative prompts and based on which we run Stable Diffusion and our method to generate corresponding images, as shown in Figure~\ref{fig:sd_vs_ours}. These examples demonstrate that the proposed layout planning and relation-aware interaction methods are able to improve the generation quality, especially in textual faithfulness.

\begin{figure*}[t]
	\includegraphics[width=0.85\textwidth]{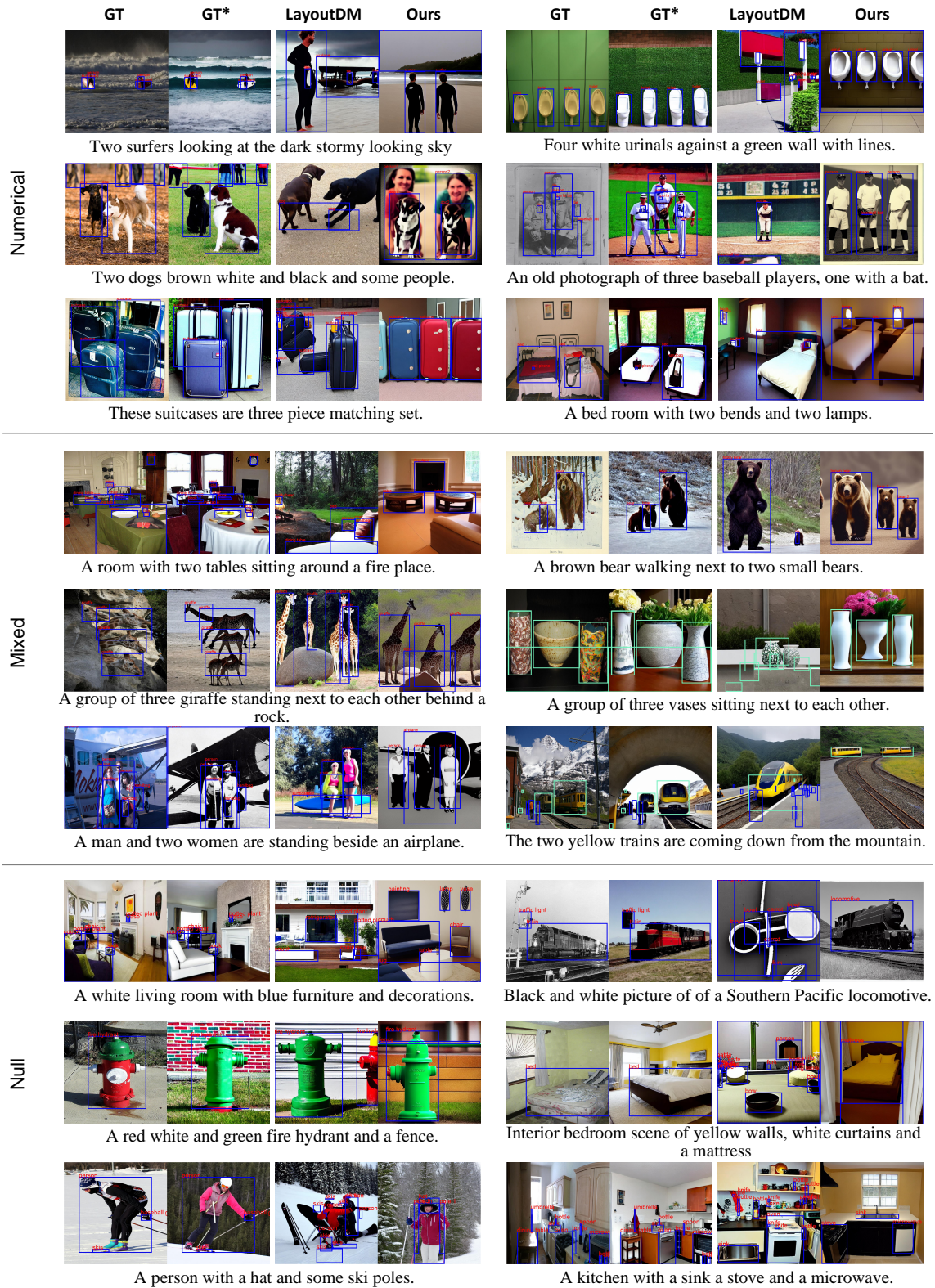}
	\caption{Qualitative results on the Numerical, Mixed, and Null test set are shown from top to bottom. The ground-truth (GT), the ground-truth layout with the generated image (GT*), the result generated LayoutDM, and our result for each prompt are shown from left to right.}
\label{fig:case2}
\vspace{-2mm}
\end{figure*}

\begin{figure*}[t]
	\includegraphics[width=0.85\textwidth]{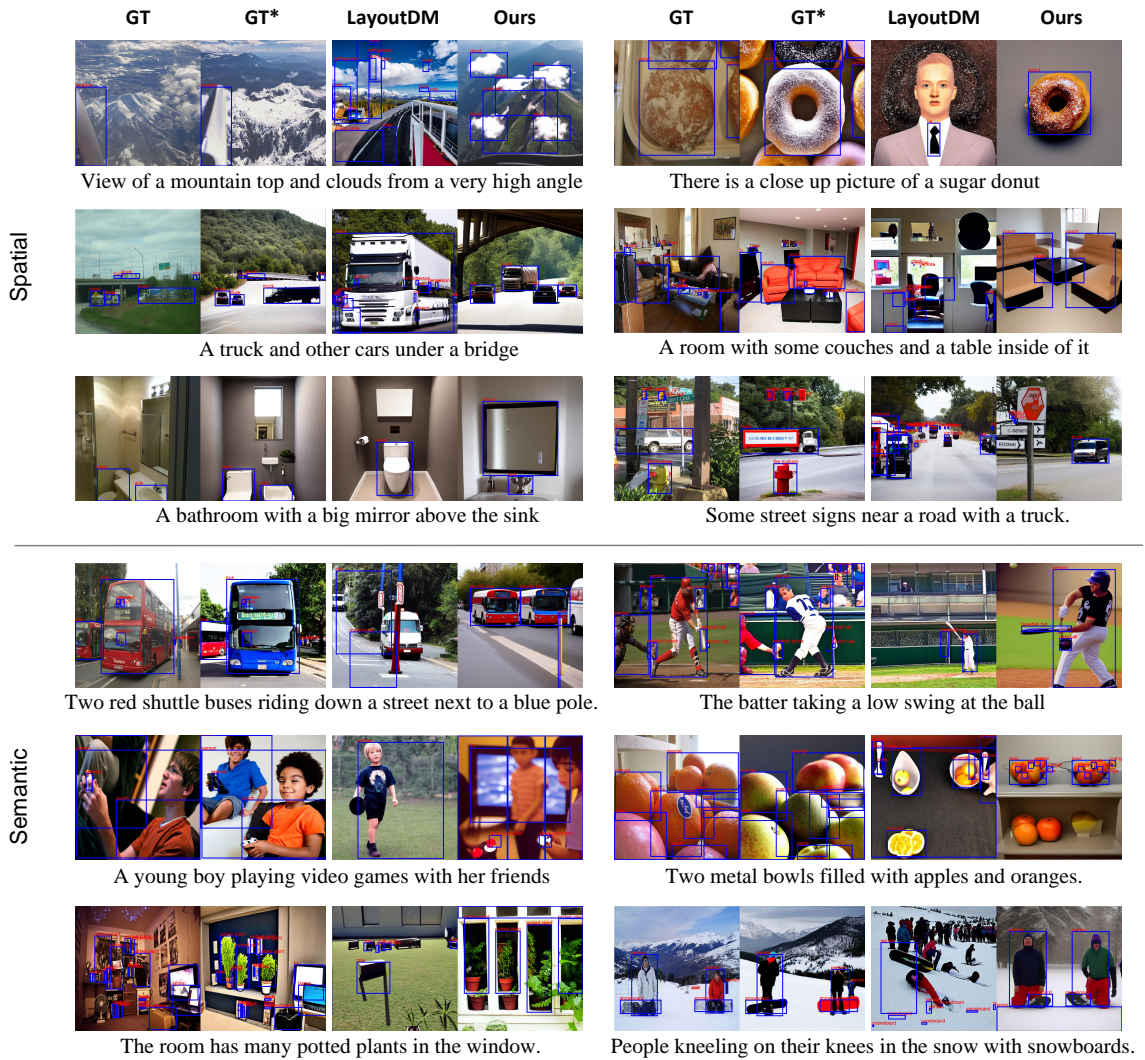}
	\vspace{-3mm}
	\caption{Qualitative results on the Spatial, and Semantic test set are depicted from top to bottom. The ground-truth (GT), the ground-truth layout with the generated image (GT*), the result generated LayoutDM, and our result for each prompt are shown from left to right.}
\label{fig:case3}
\vspace{-2mm}
\end{figure*}

\mysubsubsec{Layout-guided Generation Baselines vs. Ours.}
We provide more example images synthesized by our method and baselines in Figure \ref{fig:case2} and Figure \ref{fig:case3}. 
The results are consistent with Figure \ref{fig:case}, where our method generates images with high numerical, semantic, and spatial fidelities.

\end{document}